\PassOptionsToPackage{table}{xcolor}
\documentclass[letterpaper,twocolumn,10pt]{article}
\usepackage{hegroup}
\usepackage[table]{xcolor}
\usepackage[normalem]{ulem}
\useunder{\uline}{\ul}{}
\usepackage{tikz}
\usepackage{amsfonts}
\usepackage{xspace} 
\usepackage{amsmath}
\usepackage{mathrsfs}
\usepackage{amssymb}
\usepackage{enumitem}
\usepackage{amsmath}
\usepackage{amsthm}
\usepackage{multirow}
\usepackage{makecell}
\usepackage{caption}
\usepackage{subcaption}
\usepackage{float}
\usepackage{booktabs}
\usepackage{balance}
\usepackage{tcolorbox}
\usepackage{xurl}
\usepackage{algorithm}
\usepackage{algorithmic}
\usepackage{hyperref}
\usepackage{tabularx}
\usepackage{cleveref}

\newcommand{\method}{\textbf{\textit{ZPD-SCA}}\xspace}
\newcommand{\task}{\textbf{\textit{MSCAAE}}\xspace}
\newcommand{\mypara}[1]{\noindent{\bf {#1}.}}

\title{\method: Unveiling the Blind Spots of LLMs in Assessing Students' Cognitive Abilities}
\author{
Wenhan Dong\textsuperscript{1,}\textsuperscript{2} \ \ \ 
Zhen Sun\textsuperscript{2}  \ \ 
Yuemeng Zhao\textsuperscript{2}  \ \
Zifan Peng\textsuperscript{2}  \ \ \\ 
Jun Wu\textsuperscript{1} \ \ 
Jingyi Zheng\textsuperscript{2}  \ \ 
Yule Liu\textsuperscript{2}  \ \ \
Xinlei He\textsuperscript{2}\thanks{Corresponding author (\href{mailto:xinleihe@hkust-gz.edu.cn}{xinleihe@hkust-gz.edu.cn},~\href{mailto:molei@m.scnu.edu.cn}{molei@m.scnu.edu.cn})} \ \ \\
Yu Wang\textsuperscript{3} \ \
Ruiming Wang\textsuperscript{1} \ \
Xinyi Huang\textsuperscript{4} \ \ \
Lei Mo\textsuperscript{1}\footnotemark[1] \ \
\\
\textsuperscript{1}\textit{School of Psychology, South China Normal University} \ \ \ 
\\
\textsuperscript{2}\textit{Information Hub, Hong Kong University of Science and Technology (Guangzhou)} \ \ \ 
\\
\textsuperscript{3}\textit{School of AI, Guangzhou University} \ \ \ 
\\
\textsuperscript{4}\textit{College of Cyber Security, Jinan University} \ \ \ 
\\
}
\date{}

\begin{document}
\maketitle
\begin{abstract}
Large language models (LLMs) have demonstrated potential in educational applications, yet their capacity to accurately assess the cognitive alignment of reading materials with students' developmental stages remains insufficiently explored. 
This gap is particularly critical given the foundational educational principle of the Zone of Proximal Development (ZPD), which emphasizes the need to match learning resources with Students' Cognitive Abilities~(SCA).
Despite the importance of this alignment, there is a notable absence of comprehensive studies investigating LLMs' ability to evaluate reading comprehension difficulty across different student age groups, especially in the context of Chinese language education.
To fill this gap, we introduce~\method, a novel benchmark specifically designed to assess stage-level Chinese reading comprehension difficulty.
The benchmark is annotated by 60 ``Special Grade'' teachers, a group that represents the top 0.15\% of all in-service teachers nationwide.\footnote{The dataset construction cost at least \$20,000 (converted from RMB 150,000), primarily covering expert teacher involvement and text acquisition. 
It will be made publicly available upon acceptance of the manuscript.}
Experimental results reveal that LLMs perform poorly in zero-shot learning scenarios, with Qwen-max and GLM even falling below the probability of random guessing.
When provided with in-context examples, LLMs performance improves substantially, with some models achieving nearly double the accuracy of their zero-shot baselines.
These results reveal that LLMs possess emerging abilities to assess reading difficulty, while also exposing limitations in their current training for educationally aligned judgment.
Notably, even the best-performing models display systematic directional biases, suggesting difficulties in accurately aligning material difficulty with SCA.
Furthermore, significant variations in model performance across different genres underscore the complexity of task.
We envision that~\method can provide a foundation for evaluating and improving LLMs in cognitively aligned educational applications.
\end{abstract}

\section{Introduction}
In recent years, Large Language Models (LLMs) have made remarkable progress in educational applications~\cite{DBLP:journals/corr/abs-2403-18105,DBLP:journals/corr/abs-2405-13001,DBLP:journals/corr/abs-2505-00049}.
This progress spans a wide range of tasks, including automated essay scoring~\cite{DBLP:journals/corr/abs-2504-05736,DBLP:conf/lak/XiaoMSXZWF25} and instructional content design~\cite{DBLP:journals/corr/abs-2504-05370,DBLP:journals/corr/abs-2306-01006}.
LLMs demonstrate strong text-processing capabilities, offering new technical support for educational practices. 
However, in educational practice, providing high-quality education requires teachers to thoroughly analyze and understand students' abilities~\cite{ministry2022compulsory, Schneider01072013,LUI20161}. 
This has led to a strong assumption in the field of LLMs applied to education: \textbf{that LLMs are aware of the differences in Students' Cognitive Abilities~(SCA) across different educational stages.} 
This assumption is grounded in Vygotsky's Zone of Proximal Development~(ZPD) theory~\cite{vygotsky1978mind}, which emphasizes that education should provide learning materials aligned with students' current cognitive levels while introducing appropriate challenges to facilitate growth.

In subjects like mathematics, the requirements for students at different educational stages are often relatively objective. 
For example, the Compulsory Education Mathematics Curriculum Standard~(2022 Edition)~\cite{ministry2022MathematicsCompulsory}, outline that students should learn concepts such as fractions and inequalities. 
Specifically, second-grade students are expected to master addition and subtraction within 20~\cite{ministry2022MathematicsCompulsory}. 
In contrast, the Chinese Curriculum Standards for Compulsory Education~(2022 Edition)~\cite{ministry2022compulsory} focus on cultivating students' abilities in character recognition, reading comprehension, and writing. 
For second-grade students, the emphasis is on enjoying reading, being able to briefly retell stories, and achieving an extracurricular reading volume of no less than 50,000 words~\cite{ministry2022compulsory}. 
In practice, testing whether students have met their annual reading volume target is far more challenging than assessing their ability to perform addition and subtraction within 20. 
At the same time, the assessment of mathematical ability focuses on the mastery of specific skills, while the measurement of reading volume reflects the cumulative results of practice, representing an essential difference in assessment nature.

We observe that the strong assumption, i.e., LLMs inherently understand the SCA differences across educational stages, has been widely accepted without sufficient investigation, particularly in the domain of language skills.
This leads us to a critical question: \textbf{To what extent do LLMs accurately comprehend the SCA, especially the reading abilities of students at different educational stages?}

\begin{figure}
    \centering
    \includegraphics[width=0.45\textwidth]{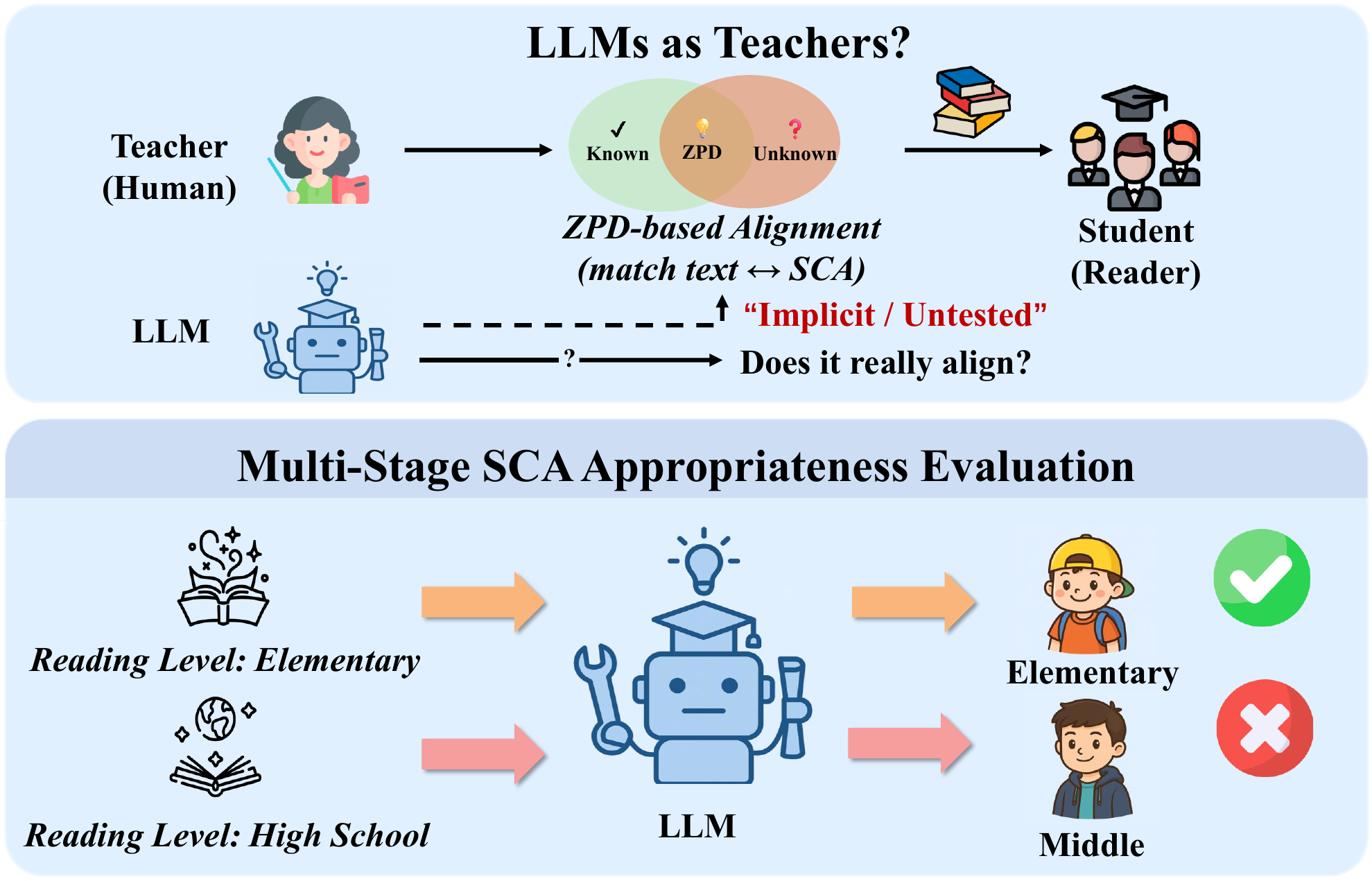}
    \caption{The motivation of our work.}
\end{figure}

This question is complex and subjective. 
Beyond examining traditional linguistic features addressed in readability assessments—such as syntactic complexity, lexical difficulty, and morphological characteristics~\cite{crossley2019moving,crossley2008assessing,collins2005predicting}, it is also necessary to incorporate analyses of content depth, logical reasoning demands, and thematic emotional complexity. 
These dimensions provide a more comprehensive reflection of the actual SCA at different educational stages.

To examine LLMs' ability to evaluate the developmental appropriateness of texts for students at different stages, we concentrate specifically on the Chinese language to examine how well LLMs understand SCA. 
The complexity of the Chinese language arises not only from its semantic richness but also from features such as multi-layered emotional connotations and the presence of words with multiple meanings that vary by context. 
These characteristics make the task of aligning texts with students' cognitive levels a significant challenge, even for experienced educators specializing in Chinese language teaching. 
In response, we define a specialized task and construct a novel benchmark, \method, designed to rigorously assess LLMs' capacity to classify Chinese texts according to their suitability for elementary, middle, and high school students.

\method is annotated by 60 ``Special Grade'' teachers, a group representing the top 0.15\% of all in-service teachers nationwide. This dataset is designed to assess LLMs' ability to classify the difficulty of Chinese texts into three educational stages: elementary, middle, and high school. 
In this task, LLMs analyze the linguistic, logical, thematic, and emotional complexity of a given text, directly mapping it to the most suitable educational stage based on an understanding of the reading abilities associated with each stage.

This study not only addresses a significant gap in the evaluation of LLMs' capabilities but also provides insights into their potential applications in education.

Our main contributions are as follows:
\begin{enumerate}
    \item To evaluate LLMs' capacity in assessing the cognitive alignment of reading materials with students' developmental stages, we frame the task as a three-way classification problem, in which LLMs are required to assign texts to one of three educational levels.
    This task evaluates whether LLMs can recognize the linguistic, logical, thematic, and emotional complexity of reading materials and align them with the cognitive capacities required for students at different developmental stages.
    This addresses a previously underexplored aspect of evaluating LLMs in education, specifically focusing on their ability to assess SCA in the context of reading comprehension.
    \item We introduce a novel benchmark~\method, which is annotated by 60 ``Special-Grade'' teachers, a nationally recognized group that represents the top 0.15\% of all K-12 educators in China, ensuring high precision and reliability. This authoritative annotated benchmark offers robust support for evaluating LLMs' ability to assess SCA in reading comprehension tasks.
    \item We reveal significant limitations of LLMs in assessing reading comprehension across educational stages, with performance variations across models. 
    These findings highlight gaps in current LLMs' ability to align text difficulty with students' developmental stages.

    \item We show that LLMs perform poorly in zero-shot settings but improve significantly with in-context learning, indicating the need for task-specific training rather than inherent capability issues. 
    By introducing~\method, we emphasize the importance of testing assumptions about LLMs' understanding of SCA, encouraging further research in educational LLMs.
\end{enumerate}

\begin{figure*}
    \centering
    \begin{subfigure}[b]{0.33\textwidth}
        \includegraphics[width=\textwidth]{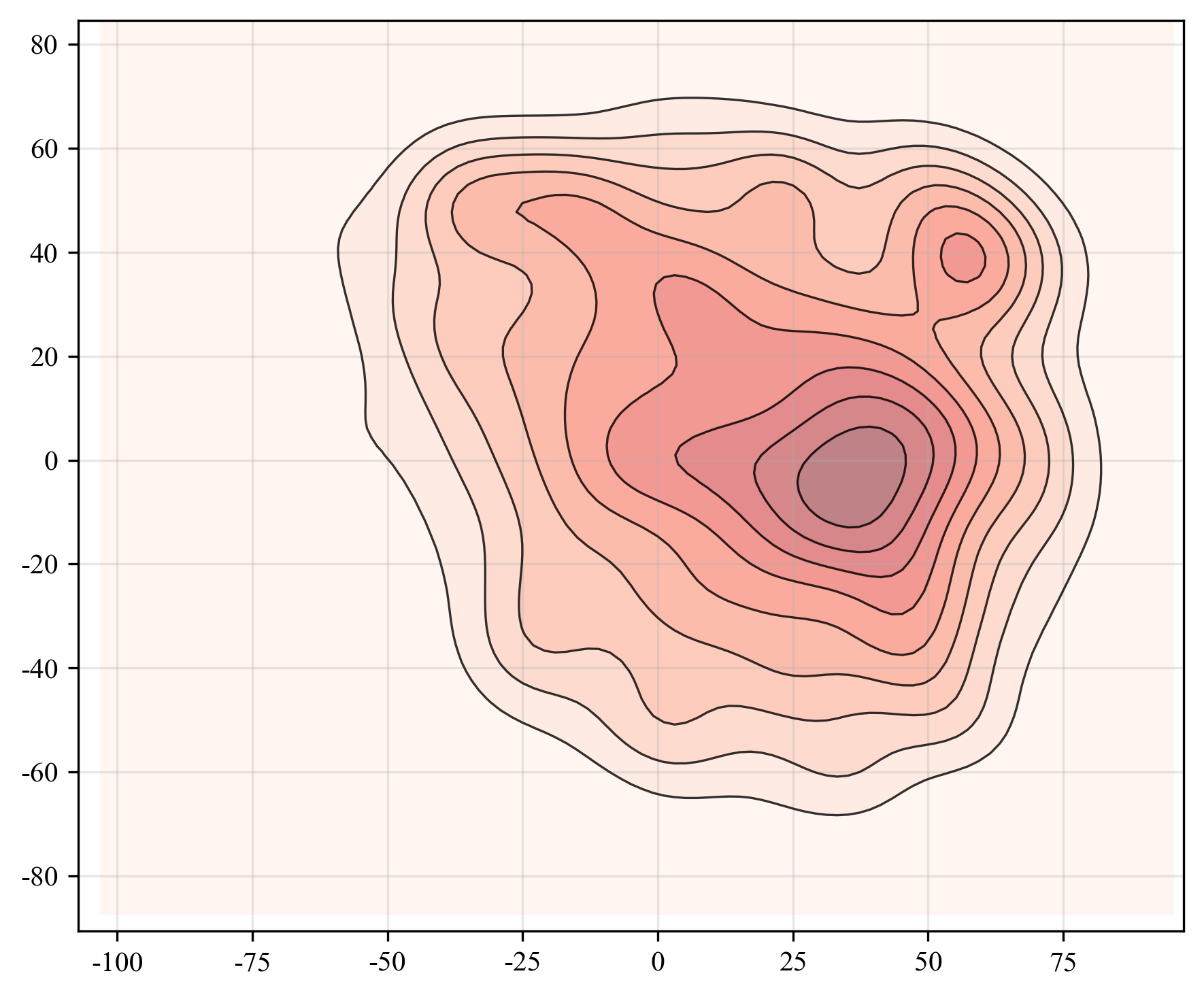}
        \caption{Elementary school}
        \label{fig:left}
    \end{subfigure}
    \hfill
    \begin{subfigure}[b]{0.33\textwidth}
        \includegraphics[width=\textwidth]{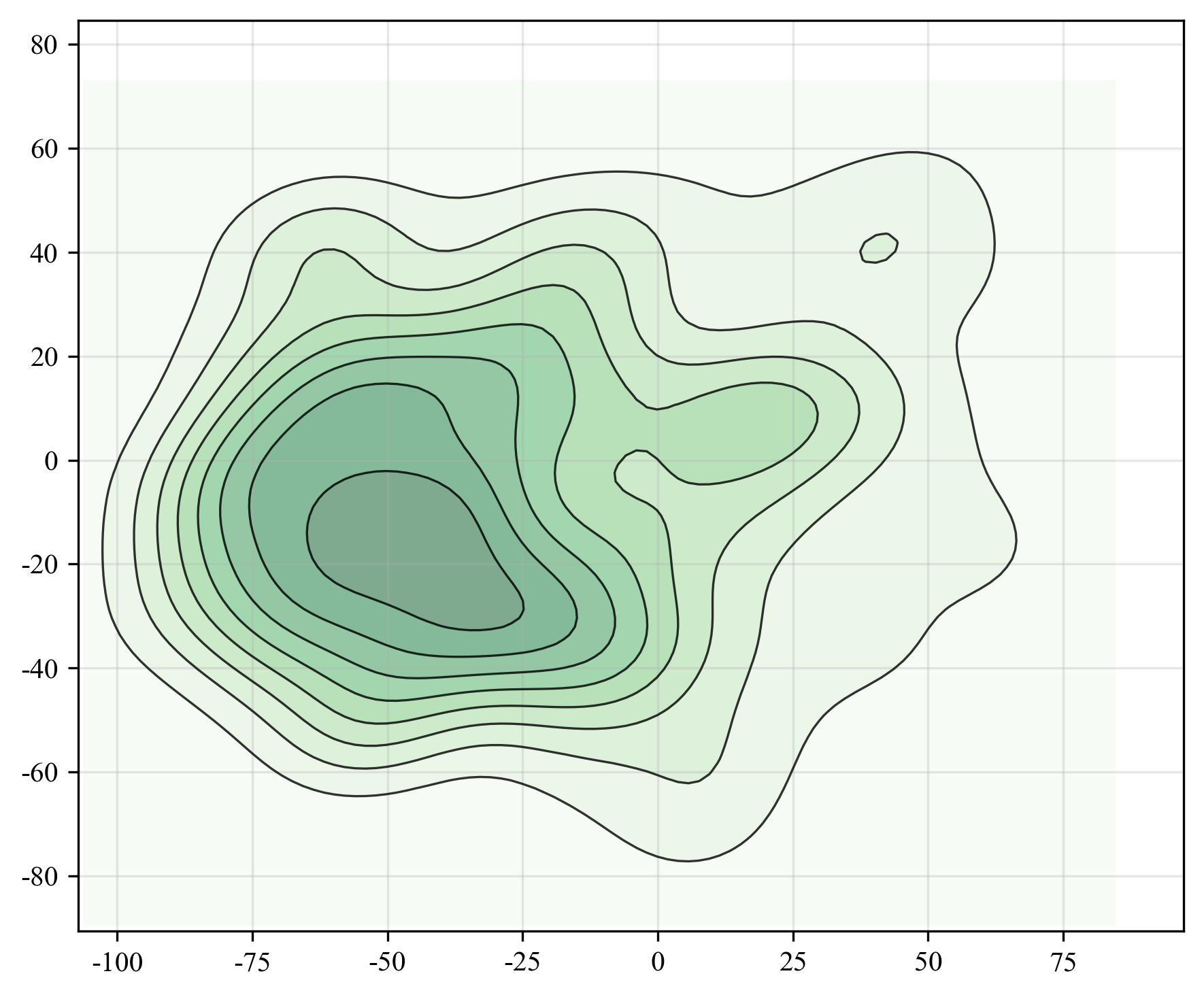}
        \caption{Middle school}
    \end{subfigure}
    \hfill
    \begin{subfigure}[b]{0.33\textwidth}
        \includegraphics[width=\textwidth]{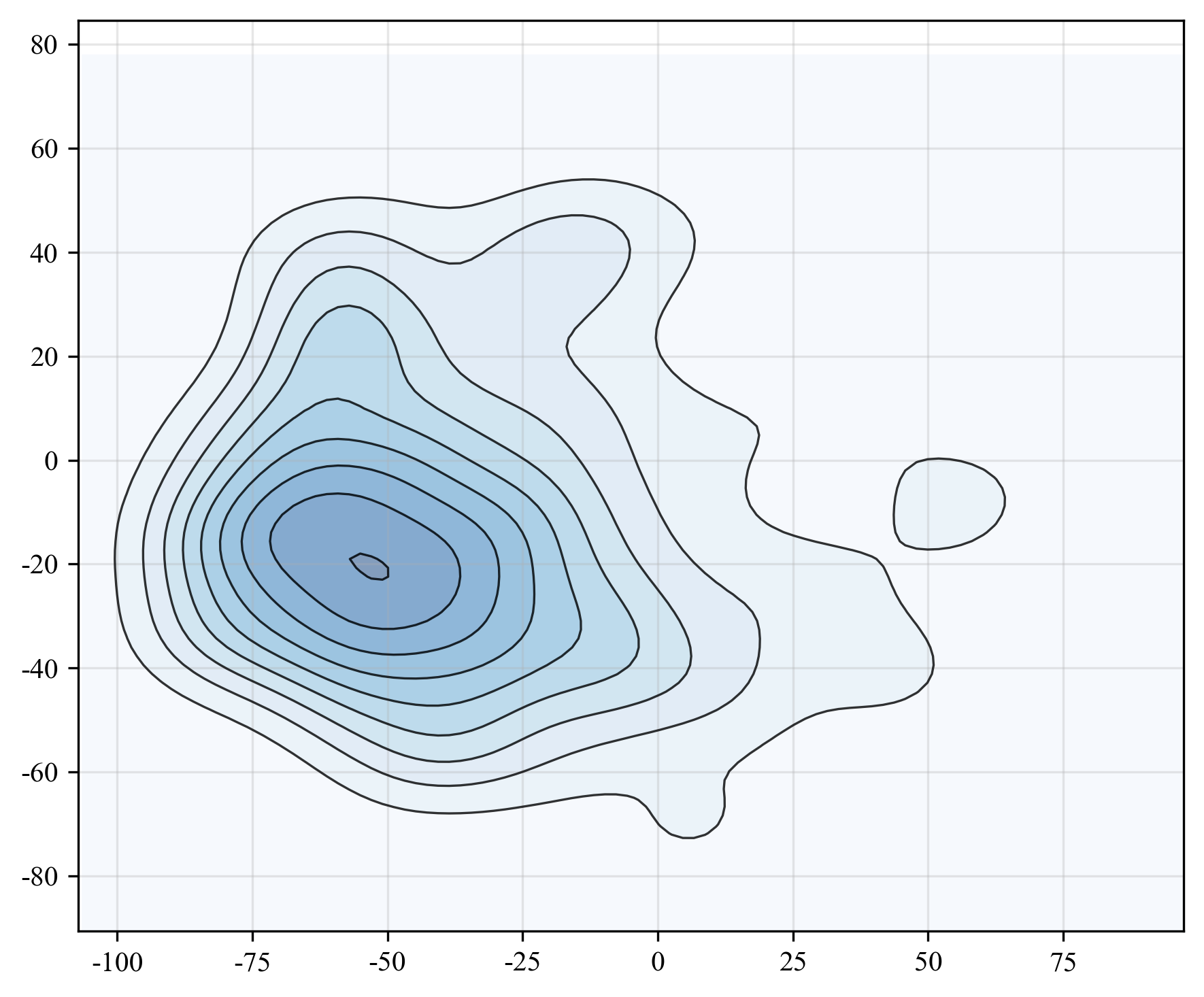}
        \caption{High school}
    \end{subfigure}
    \caption{Semantic space analysis of educational stages. 
    This figure presents the t-SNE contour visualizations of text embeddings for three educational stages: elementary, middle, and high school. 
    The results indicate that, from a purely semantic perspective, elementary school exhibits a more distinct distribution in the embedding space, while middle school and high school demonstrate greater overlap, suggesting higher difficulty in distinguishing between these two categories based solely on semantic features.}
    \label{fig:combined}
\end{figure*}

\section{Task Definition}
\subsection{Multi-Stage SCA Appropriateness Evaluation~(\task)}
This section introduces a new task~\task, designed to evaluate the ability of LLMs to determine the reading comprehension difficulty of Chinese texts. 
The primary objective of this task is to assess whether these models can accurately classify texts into one of three educational stages-elementary, middle, or high school-within an annotated dataset, while capturing the cognitive differences in Chinese reading comprehension abilities across these stages.

To accomplish~\task, the input texts need to satisfy two essential criteria.

First, the content should come from real-world reading materials rather than standardized textbooks or curricula. 
These materials may include extracurricular readings, news articles, and popular science literature, ensuring alignment with real-world reading scenarios. 

Second, the text difficulty should match the average Chinese reading comprehension level of students at a specific educational stage. 
The difficulty is generally assessed based on the following dimensions: 
linguistic complexity, thematic depth, logical reasoning, and emotional complexity.

For each input text, the model is required to generate a classification label that identifies the most appropriate educational stage. 
For example, ``Elementary'' indicates that the text is suitable for elementary school students, ``Middle'' for middle school students, and ``High'' for high school students. 
The classification accuracy directly influences the model's potential applications in Chinese education, especially in personalized learning resource recommendations and reading ability evaluations.
Importantly, the task focuses on evaluating the model's ability to cognitively assess students' reading comprehension capabilities at different educational stages by analyzing the alignment between text difficulty and students' cognitive development.
Thus, the model must not only identify text difficulty but also integrate the cognitive development traits of students at various educational stages to comprehensively judge its understanding of student abilities.
Specifically, the model needs to possess the following two core capabilities:
\begin{itemize}
    \item \textbf{Understanding Students' Chinese Reading Abilities at Specific Educational Stages}: The model must grasp the cognitive characteristics of elementary, middle, and high school students in terms of language comprehension, logical reasoning, knowledge background, and thematic acceptance. 
    \item \textbf{Evaluating the Reading Difficulty of Chinese Texts}: The model must comprehensively analyze text difficulty across linguistic complexity, thematic depth, logical reasoning, and emotional complexity, aligning it with students' reading abilities. 
\end{itemize}

\subsection{Evaluation Metrics}
To assess the model's ability to classify Chinese reading comprehension difficulty levels, we propose two evaluation metrics: Cross-Level Migration Concentration (CLME) and Weighted Directional Bias Index (WDBI).

\mypara{CLME}
The CLME measures the proportion of misclassifications occurring between non-adjacent difficulty levels.
This metric $M_{ij}$ denotes the number of samples classified as class $j$ while their true class is $i$. The errors are weighted by the distance between classes to indicate the severity of misclassification.
The CLME is defined as:

\begin{equation}
\text{CLME} = \frac{\sum_{i,j, |i - j| > 1} M_{ij}}{\sum_{i,j, i \neq j} M_{ij}} \notag
\end{equation}

\mypara{WDBI}
The WDBI evaluates the model's overall tendency to overestimate or underestimate difficulty levels, while adjusting for class imbalance. 
It compares the weighted counts of upward shifts (predicting higher difficulty than the true label) and downward shifts (predicting lower difficulty than the true label).
The WDBI is defined as:

\begin{equation}
\text{WDBI} = \frac{\sum_{i < j} w_i M_{ij} - \sum_{i > j} w_i M_{ij}}{\sum_{i,j} w_i M_{ij}} \notag
\end{equation}

\begin{figure*}
    \centering
    \includegraphics[width=\textwidth]{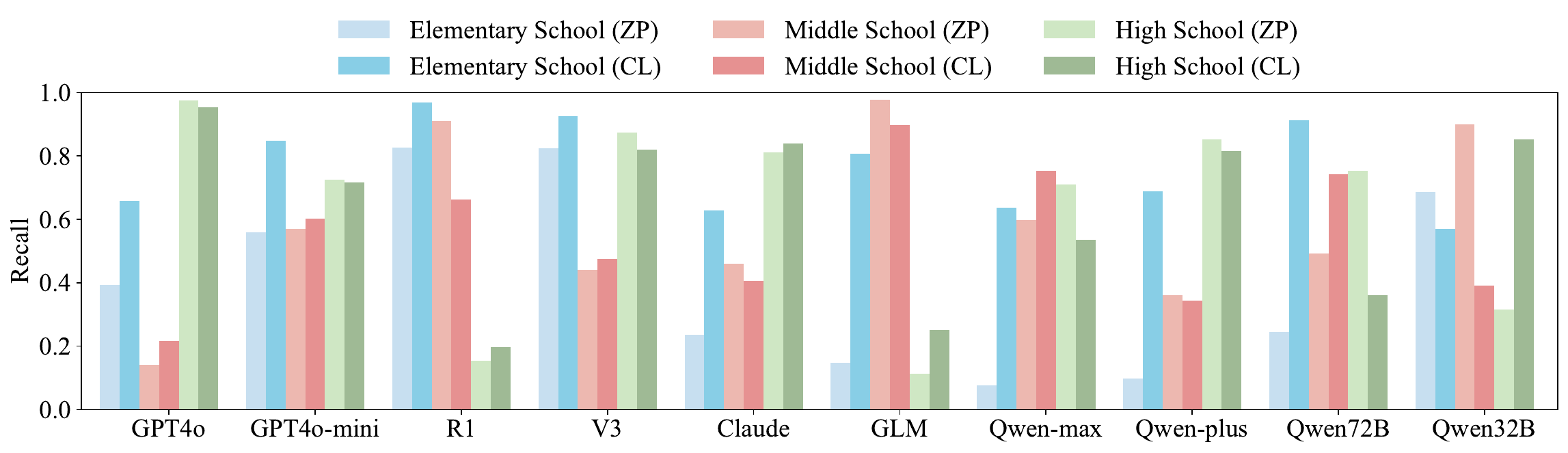} 
    \caption{Recall rates for elementary, middle, and high school difficulty levels across LLMs. 
    The figure illustrates how well each model identifies texts belonging to difficulty categories, highlighting variations in recall performance across labels.} 
    \label{fig:recall_plot} 
\end{figure*}

\section{\method}
The data collection process was carefully structured to encompass both diversity and educational appropriateness. 
Researchers with expertise in educational psychology collected raw data by selecting texts from a diverse range of extracurricular books, ensuring suitability for elementary, middle and high school students.
To ensure comprehensive coverage of reading materials typically encountered by students, we incorporated a wide range of genres, including fairy tales, fantasy, science fiction, campus life, adventure, and others, totaling \textbf{12 distinct categories}.
These genres not only align with the typical categories in students' extracurricular reading but also capture the varying levels of reading comprehension required at different educational stages.

This rigorous selection process resulted in a corpus of over \textbf{8 million characters}, providing a robust foundation for subsequent analysis.
During the data annotation phase, we collaborated with 60 distinguished educators, all of whom were recipients of the title ``Special Grade'' teachers. 
This prestigious title is conferred by the Chinese government upon top-performing teachers, with the number of awardees not exceeding 0.15\% of all active teachers nationwide. 

Each book was independently evaluated by at least 20 annotators, who assessed its suitability for specific educational stages.
All teachers participating in the annotation work were compensated at the highest standard permitted by the state for their professional services.

After annotation and filtering, the dataset was reduced from an initial 8 million characters to a high-quality labeled corpus of \textbf{4.7 million characters}. 
This reduction resulted from the exclusion of unsuitable or controversial texts, thereby ensuring the consistency and reliability.

\mypara{Semantic exploration of \method}
We explored the semantic differences in the annotated samples using GTE-large~\cite{li2023generaltextembeddingsmultistage}.
After obtaining the embeddings, we applied t-SNE to visualize and analyze the data. 
As shown in~\Cref{fig:combined}, middle and high school samples are not easily distinguishable at the semantic level. 
However, elementary samples show clear differences from both middle and high school samples. 
Clustering analysis confirms this, achieving 81.2\% accuracy in separating elementary samples from the others.

\section{Experiments and Analysis}
This section outlines the experimental setup, results, and analysis conducted to evaluate LLMs in assessing the difficulty stages of Chinese reading comprehension texts for three stages: elementary, middle, and high school. 
\subsection{Experimental Setup}
We evaluate 10 LLMs, including GPT4o (version 2024-08-06), GPT4o-mini (version 2024-07-18)~\cite{DBLP:journals/corr/abs-2303-08774}, Deepseek-R1 (abbreviated as R1), Deepseek-V3 (abbreviated as V3)~\cite{DBLP:journals/corr/abs-2412-19437}, Claude3.5-haiku (abbreviated as Claude)~\cite{anthropic_introducing_claude}, GLM-4-plus (abbreviated as GLM)~\cite{zhipu_bigmodel_introduction}, Qwen-max (version 2024-10-15), Qwen-plus (version 2025-06-24), Qwen2.5-72b-instruct (abbreviated as Qwen72B), and Qwen2.5-32b-instruct (abbreviated as Qwen32B)~\cite{DBLP:journals/corr/abs-2412-15115}. 

\begin{table*}[ht]
\setlength{\tabcolsep}{6pt} 
\renewcommand{\arraystretch}{0.95} 
\centering
\caption{Comparison of ACC scores achieved by the evaluated models under ZP and CL settings. 
These scores indicate how much better the models perform in CL compared to ZP in terms of ACC.}
\label{tab:acc_diff}
\resizebox{0.98\textwidth}{!}{
\begin{tabular}{c|c|c|c|c|c|c|c|c|c|c}
\toprule
Model$\rightarrow$ & GPT4o & GPT4o-mini & R1 & V3 & Claude & GLM & Qwen-max & Qwen-plus & Qwen72B & Qwen32B \\
\midrule
All& $+$0.1676 & $+$0.1786 & $+$0.0578 & $+$0.0538 & $+$0.2356 & $+$0.4219 & $+$0.3235 & $+$0.3522 & $+$0.2065 & $+$0.1241 \\
Elementary&$+$0.2657 &$+$0.2895 &$+$0.1410 &$+$0.1007 &$+$0.3909 &$+$0.6607 &$+$0.5587 &$+$0.5901 &$+$0.3247 &$+$0.2267 \\
Middle&$+$0.0765 &$+$0.0314 & $-$0.2481 &$+$0.0335 & $-$0.0527 & $-$0.0812 &$+$0.1559 & $-$0.0171 & $-$0.1005 & $-$0.1568 \\
High& $-$0.0227 & $-$0.0079 &$+$0.0432 & $-$0.0534 &$+$0.0266 &$+$0.1364 & $-$0.1753 & $-$0.0352 &$+$0.0978 &$+$0.0446 \\
\bottomrule
\end{tabular}}
\end{table*}

\begin{table*}
\setlength{\tabcolsep}{6pt} 
\renewcommand{\arraystretch}{0.95} 
\centering
\caption{Performance evaluation of 10 different models across various text genres in CL. 
``All'' represents the aggregate performance across all genres. 
F1 and ACC indicate the model's performance for a specific genre. 
CLME assesses the proportion of misclassifications occurring across non-adjacent difficulty levels, while WDBI evaluates the overall tendency of the model to overestimate or underestimate difficulty levels, accounting for class imbalance. For a complete breakdown of results across individual genres and ZP performance, see Appendix~\Cref{tab:big} and~\Cref{tab:big_zero}.}
\label{tab:small_CL}
\resizebox{0.98\textwidth}{!}{
\begin{tabular}{c|c|c|c|c|c|c|c|c|c|c|c}
\toprule
Dataset&\makecell{Model$\rightarrow$\\ Metrics $\downarrow$}& GPT4o &GPT4o-mini&R1&V3&Claude&GLM&Qwen-max&Qwen-plus&Qwen72B&Qwen32B\\
\midrule
\multirow{4}{*}{All}
&\makecell{F1} & 0.6896 & 0.7993 & 0.7184 & 0.8352 & 0.6860 & 0.7038 & 0.6818 & 0.6962 & 0.6547 & 0.7610 \\
&\makecell{ACC} & 0.6618 & 0.7784 & 0.7413 & 0.8313 & 0.6445 & 0.6878 & 0.6296 & 0.6674 & 0.6091 & 0.7555 \\
&\makecell{CLME} & 0.0477 & 0.0189 & 0.0874 & 0.0350 & 0.0329 & 0.0074 & 0.0062 & 0.0924 & 0.0361 & 0.0396 \\
&\makecell{WDBI} & 0.3561 & 0.0704 & -0.3243 & 0.1081 & 0.2579 & -0.1708 & 0.3727 & 0.1684 & 0.2953 & -0.1930 \\
\midrule
\multirow{4}{*}{Academic}
&\makecell{F1} & 0.9873 & 0.7500 & 0.2222 & 0.8571 & 0.9744 & 0.3750 & 0.7302 & 0.9333 & 0.9474 & 0.5185 \\
&\makecell{ACC} & 0.9750 & 0.6000 & 0.1250 & 0.7500 & 0.9500 & 0.2308 & 0.5750 & 0.8750 & 0.9000 & 0.3500 \\
&\makecell{CLME} & 0.0000 & 0.0000 & 0.0000 & 0.0000 & 0.0000 & 0.0000 & 0.0000 & 0.6000 & 0.0000 & 0.0000 \\
&\makecell{WDBI} & -0.0250 & -0.4000 & -0.8750 & -0.2500 & -0.0500 & -0.7692 & 0.4250 & -0.1250 & -0.1000 & -0.6500 \\
\midrule
\multirow{4}{*}{Fantasy}
&\makecell{F1} & 0.4273 & 0.7598 & 0.9294 & 0.9166 & 0.4379 & 0.8748 & 0.5911 & 0.5993 & 0.4745 & 0.9427 \\
&\makecell{ACC} & 0.2922 & 0.6712 & 0.9296 & 0.8904 & 0.3178 & 0.8356 & 0.4749 & 0.4717 & 0.3288 & 0.9315 \\
&\makecell{CLME} & 0.0258 & 0.0139 & 0.0000 & 0.0417 & 0.0068 & 0.0000 & 0.0087 & 0.0268 & 0.0408 & 0.0000 \\
&\makecell{WDBI} & 0.7005 & 0.2108 & -0.2474 & 0.1303 & 0.5431 & 0.0465 & 0.5297 & 0.3842 & 0.7171 & 0.0364 \\
\bottomrule
\end{tabular}}
\end{table*}

Their performance is evaluated using four metrics: F1 score, accuracy (ACC), CLME, and WDBI.
While F1 and ACC capture overall classification accuracy, CLME measures the severity of errors by penalizing non-adjacent misclassifications, and WDBI evaluates the directional bias in predictions (e.g., whether models tend to overestimate or underestimate SCA when aligning texts with appropriate difficulty levels).

We consider two templates for prompting the models:
\begin{itemize}
\item \textbf{Zero-shot Prompting (ZP)}: The LLMs are provided with a brief description of the criteria and the full text as input, directly outputting a three-class result.
\item \textbf{In-Context Learning (CL)}: In addition to using the same input and output format as zero-shot prompting, illustrative examples are included in the prompt.
\end{itemize}

\subsection{Prompt Design}
The prompts used in our experiments are developed through an iterative refinement process, leveraging the capabilities of GPT-4o and Qwen-max to provide structured feedback. 
Initially, we test basic prompts on these models, which are designed to elicit three-class difficulty level predictions corresponding to elementary, middle, and high school cognitive abilities. 
After collecting sample outputs, we systematically identify instances of misclassification and analyze the reasoning provided by the models in conjunction with the ground truth labels. 
This analysis is then used as input for further iterations, where GPT-4o and Qwen-max are tasked with refining the prompts to enhance clarity and consistency. 
Through this iterative process, we ensure that the final prompts provide sufficient contextual information. 

It is important to note that the objective of this process is not to optimize prompts for maximum performance but rather to develop a consistent and reasonable framework for testing model capabilities across different settings.

\subsection{Experimental Results}
\Cref{tab:acc_diff} compares the ACC achieved by the evaluated models under ZP and CL settings, highlighting the impact of contextual examples on performance. 
\Cref{tab:small_CL} summarizes the key performance metrics (F1, ACC, CLME, and WDBI) for some genres under the CL condition.

\mypara{Initial underperformance in zero-shot prompting}
In the ZP, several models exhibit significant challenges in accurately classifying text difficulty levels. 
For example, GLM and Qwen-max achieve accuracy scores below 33\%, which is lower than the expected performance of random guessing in a three-class classification task.
In contrast, Qwen32B performs relatively better, achieving an ACC of 63.14\%. 
This discrepancy suggests that these models may lack sufficient training in recognizing the SCA at different educational stages.
Furthermore, models with strong general text comprehension capabilities might inadvertently misinterpret SCA levels due to their extensive exposure to diverse but non-targeted training data, potentially introducing biases into their predictions.

Notably, while elementary school samples can be distinguished with (81.2\%) accuracy semantically, models like Qwen-max and GLM show a stark contrast in the ZP setting. 
Their ACC for elementary school samples (9.83\% and 7.71\%, respectively) falls significantly below the random guessing baseline. 
This highlights a critical gap: \textbf{while the models can differentiate text difficulty semantically, they fail to effectively recognize SCA}.

Further analysis of GLM shows that its CLME score was 0, suggesting an absence of severe misclassifications across non-adjacent difficulty levels.
However, this outcome arises from the model's overwhelming tendency to assign nearly all texts (88.03\%) to the middle school category, revealing a fundamental inability to differentiate between difficulty levels effectively. 
Additionally, among the 10 models, seven exhibit positive WDBI values, with an absolute average WDBI of 0.2696 and a mean WDBI of 0.2035.
Notably, Qwen-Plus and GPT-4o had WDBI values exceeding 0.4, suggesting a pronounced tendency to underestimate human cognitive abilities and overestimate text difficulty.

\mypara{Unlocking potential with few-shot learning}
When evaluated under the CL, the performance of these models improved dramatically. 
For example, Qwen-max's ACC increases from 30.61\% to 62.96\%, representing an improvement of over 2.06$\times$. 
Similarly, the ACC of GLM rises from 26.59\% to 68.78\%, a gain of approximately 2.59$\times$. 
These substantial improvements highlight the critical role of targeted training or contextual examples in enabling LLMs to perform well on tasks requiring nuanced understanding.
The results may suggest that initial underperformance is not due to inherent limitations in the models but rather the absence of exposure to relevant training data or task-specific contexts.

In the CL setting, we provide two elementary-level examples to help the models calibrate their understanding of human cognitive abilities. 
This adjustment reduces the absolute average WDBI by 14\%, bringing it down to 0.2317, while the mean WDBI decreased to 0.094. 
These changes suggest a partial reduction in the models' tendency to underestimate SCA, leading to a notable improvement in classification accuracy.
However, three LLMs that initially overestimate human abilities further exacerbated this bias. 
To determine whether the observed improvement is influenced by label imbalance (e.g., models favoring the elementary label to achieve higher accuracy), we examine recall rates for texts aligned with the three educational stages.
We observe improvements in recall rates for middle and high school across most models (see \Cref{fig:recall_plot}), ruling out the possibility of label bias. 
These findings reinforce our hypothesis that the LLMs possess some inherent ability to distinguish text difficulty but lacked appropriate training or context.

Under the CL, V3 achieves the highest overall F1 (83.52\%) and ACC (0.8313), demonstrating its strong capability in assigning texts to the correct difficulty levels. 
Despite this, WDBI (0.1081) indicates persistent challenges in accurately capturing the directionality of classification errors, suggesting opportunities for further refinement.

Notably, smaller models such as GPT-4o-mini and Qwen32B demonstrate strong competitiveness in terms of F1 scores and accuracy, suggesting that model size alone does not guarantee superior performance on this task.
The performance of models varied significantly across different genres, reflecting the complexity and diversity of the texts. 
In the Academic genre, Claude achieves the highest accuracy (0.9500).
Meanwhile, in the fantasy genre, Qwen32B surpasses other models with an F1 score of 0.9427 and the lowest CLME (0), demonstrating its ability to minimize severe misclassifications.

\section{Enhancing Capability through Training}

\begin{table}[h!]
\centering
\small
\caption{Comparison of Qwen-32B performance before and after LoRA fine-tuning.}
\label{tab:funetuned}
\resizebox{0.48\textwidth}{!}{
\begin{tabular}{lcccc}
\toprule
\textbf{Setting} & \textbf{Accuracy} & \textbf{F1} & \textbf{CLME} & \textbf{WDBI} \\
\midrule
Zero-Shot& 0.6314 & 0.6696 & 0.0046 & -0.1046 \\
Few-Shot   & 0.7555 & 0.7610 & 0.0396 & -0.1930 \\
\textbf{Fine-Tuned} & \textbf{0.7986} & \textbf{0.6169} & \textbf{0.0467} & \textbf{-0.0502} \\
\bottomrule
\end{tabular}
}
\end{table}

We conduct fine-tuning on the Qwen32B using Low-Rank Adaptation (LoRA)~\cite{DBLP:conf/iclr/HuSWALWWC22} under ZP setting. 
The LoRA configuration uses a rank of 32, a learning rate of $5 \times 10^{-4}$, and a batch size of 8, trained for 5 epochs.

\Cref{tab:funetuned} summarizes the evaluation results after training, compared with Qwen32B's performance under zero-shot and few-shot settings.
The fine-tuned Qwen32B achieves an accuracy of 0.7986 in a zero-shot setting, surpassing its previous few-shot performance (0.7555) and significantly improving upon the original zero-shot result (0.6314).
The CLME score increases slightly (from 0.0046 to 0.0467), but remains within an acceptable range, indicating no severe misclassification. 
Moreover, the WDBI is closer to zero (-0.0502), suggesting a reduction in directional bias compared to both prior settings. 
These results demonstrate that targeted fine-tuning via LoRA can not only bridge the gap between ZP and CL capabilities, but also lead to more balanced model behavior.

\section{Discussion}
\mypara{Zero-shot performance underestimates models' latent potential in aligning text with student cognitive levels}
Our experiments reveal that ZP performance significantly underestimates the ability of LLMs to assess the suitability of texts for different cognitive levels. 
When provided with in-context examples, model accuracy improves substantially, with some models achieving 2-3$\times$ higher accuracy compared to their ZP baselines. 
This suggests that LLMs have latent knowledge about text complexity and its alignment with SCA levels but require appropriate contextual guidance to better utilize this capability.
The marked improvement through CL indicates that these models likely lack exposure to educational tasks during pretraining, which limits their ability to perform such specialized assessments effectively.

\mypara{Model size and general leaderboard rankings do not consistently predict success in assessing text alignment with SCA}
Contrary to expectations, smaller LLMs such as Qwen32B and GPT-4o-mini outperformed larger counterparts like Qwen-max, and GPT-4o in evaluating text suitability for different SCA levels.
For instance, despite their high rankings on general benchmarks~\cite{hendrycks2020measuring,wang2024mmlu,chen2021evaluating,goyal2022flores,bandarkar2023belebele}, GPT-4o and Qwen-max struggled to accurately assess text difficulty and SCA alignment.
This finding highlights that model scale alone is insufficient for success in specialized educational tasks.

\begin{figure}
    \centering
    \includegraphics[width=0.45\textwidth]{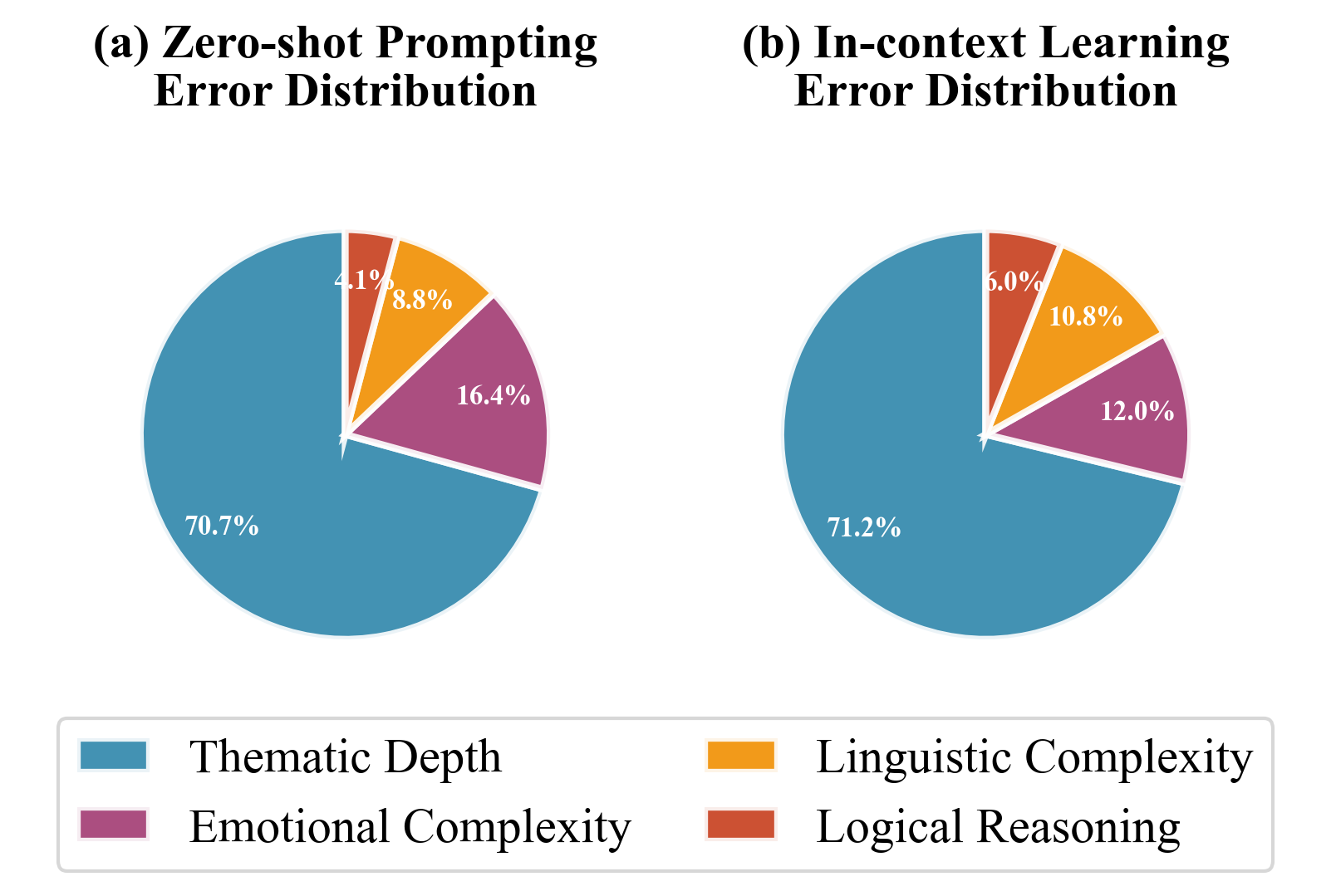}
    \caption{Comparative analysis of error type classification. This figure illustrates the misprediction categorization of ten LLMs, where the error attributions were analyzed and classified by GPT-4o based on the input of mispredictions and their underlying causes. 
    Categories accounting for less than 0.1\% are excluded from visualization for clarity.For detailed results of individual models, see Appendix~\Cref{fig:error_analysis}.}
    \label{fig:error}
\end{figure}

We hypothesize that larger models, trained on diverse datasets, may inadvertently develop biases that conflict with the nuanced requirements of educational applications focused on student cognitive levels.
Their vast parameter spaces and broad exposure to general domains might lead them to over-rely on patterns irrelevant to the task at hand.
In contrast, smaller models, constrained by their limited capacity, appear less influenced by such biases and can focus more effectively on task-specific features, such as aligning texts with cognitive levels.

\mypara{CL has a pronounced impact, particularly on larger models}
The transition from ZP to CL yielded particularly significant improvements for larger models. 
For example, GLM saw its accuracy increase by approximately 2.59$\times$ when provided with contextual examples. 
This suggests that larger models, with their greater parameter space and learned representations, have a stronger capacity to leverage in-context information effectively. 
However, smaller models like Qwen32B also demonstrated competitive performance, indicating that model size alone is not the sole determinant of success in this task. 
These results underscore the importance of targeted approaches, such as CL, to unlock the full potential of LLMs in educational contexts.

\mypara{Evidence points to a gap in training for cognitive-level alignment tasks}
Our findings collectively indicate that the suboptimal performance of LLMs in assessing text alignment with student cognitive levels likely results from insufficient exposure to relevant educational tasks during pretraining. 
While LLMs demonstrate latent capabilities that can be enhanced through CL, their current training does not adequately equip them for specialized evaluations of reading comprehension difficulty across different developmental stages. 
This limitation is particularly significant given the critical need to align educational materials with students' cognitive levels, as highlighted by ZPD theory.

In~\Cref{fig:error}, CL proves a certain level of effectiveness in aligning aspects such as ``Emotional Complexity'' and ``Linguistic Complexity.'' 
However, it falls short in addressing more intricate dimensions like ``Thematic Depth'' and ``Logical Reasoning'', where more targeted training strategies are evidently required. 
Bridging this gap will necessitate focused training approaches and further exploration into the intersection of LLMs and education. 
Overall, addressing this issue remains an open challenge and calls for continued research in this interdisciplinary domain.

\section{Related Work}
\mypara{LLMs as cognitive models of reading}
Recent studies indicate that LLMs can serve as cognitively plausible proxies for human reading processes~\cite{cevoli2022prediction,gu2025reading}.
For example, Lopes Rego et al.~\cite{rego2024cloze} demonstrated that replacing traditional cloze predictability with LLM-based predictions significantly improved a cognitive simulation of eye-movement behavior during reading.

\mypara{LLM-powered educational tools for reading comprehension}
A growing body of work leverages LLMs to generate and personalize reading materials for learners~\cite{xiao2023chatgpt,ge2024chatgpt,vorobyeva2025personalized,huynh2025genai}.
Such personalized or simplified content has been shown to enhance learners' engagement and understanding.
While these LLMs show promise in reducing teacher workload and providing adaptive reading practice, their outputs still require careful educator vetting to ensure appropriate difficulty and alignment with curricular goals.

\mypara{Limitations of LLMs in SCA assessment, particularly in Chinese}
While existing research has explored the potential of LLMs in educational applications~\cite{opesemowo2024systematic,annuvs2024learn,naz2024exploring,adel2024chatgpt}, a critical gap remains in their ability to directly assess SCA, particularly in the context of reading comprehension across different educational stages. 
Despite advancements in text processing capabilities~\cite{sanghera2025high,allaway2024exceptions,DBLP:journals/corr/abs-2505-23843,lopes2024language}, most studies have primarily focused on leveraging LLMs for tasks such as text simplification or personalized content generation, often assuming that these models inherently understand the SCA differences between elementary, middle, and high school. 
However, this assumption has not been rigorously tested, especially in languages like Chinese, which present unique challenges due to their semantic richness and contextual variability. 
Although personalized LLMs can estimate individual proficiency levels by learning from specific user data, there is also a crucial need to delineate the boundaries of group-level cognitive abilities. 
Notably, with appropriate prompting strategies, LLMs may demonstrate some ability to discern the boundaries of cognitive difficulty levels across different educational stages~\cite{xiao2025exploring,kim2023study,nguyen2023efl}; however, comprehensively understanding these broader patterns remains critical for designing equitable and effective educational tools~\cite{bal2025potential,chardonnens2025adapting,ghafouri2024virtual}. 

\section{Conclusion}
Based on Vygotsky's Zone of Proximal Development (ZPD) theory, evaluating the alignment between text difficulty and students' cognitive levels is a core issue in educational applications. 
However, current large language models (LLMs) generally lack targeted consideration for this task during their training process. 
Our study introduces a novel task framework to assess LLMs' ability to evaluate Students' Cognitive Abilities (SCA) in reading comprehension across different educational stages.
By constructing a high-quality benchmark~\method annotated by 60 ``Special Grade'' teachers, we systematically uncover the limitations of existing LLMs in performing this task. 
Experimental results reveal that while in-context learning significantly improves model performance, the low accuracy in zero-shot scenarios, along with cross-level misclassification and directional bias issues, highlights the neglect of educational-specific needs in current model training. 
These findings not only provide critical directions for enhancing the adaptability of LLMs in educational contexts but also underscore the importance of incorporating cognitive alignment tasks into model training to advance their application in education.

\bibliographystyle{plain}
\bibliography{ref}

\begin{thebibliography}{10}

\bibitem{adel2024chatgpt}
Amr Adel, Ali Ahsan, and Claire Davison.
\newblock Chatgpt promises and challenges in education: Computational and ethical perspectives.
\newblock {\em Education Sciences}, 14(8):814, 2024.

\bibitem{allaway2024exceptions}
Emily Allaway, Chandra Bhagavatula, Jena~D Hwang, Kathleen McKeown, and Sarah-Jane Leslie.
\newblock Exceptions, instantiations, and overgeneralization: Insights into how language models process generics.
\newblock {\em Computational Linguistics}, 50(4):1211--1275, 2024.

\bibitem{annuvs2024learn}
Norbert Annu{\v{s}} and Tibor Kmet'.
\newblock Learn with me—let us boost personalized learning in k-12 math education!
\newblock {\em Education Sciences}, 14(7):773, 2024.

\bibitem{anthropic_introducing_claude}
{Anthropic}.
\newblock Introducing claude.
\newblock Anthropic website, 2023.
\newblock Next‑generation AI assistant, Claude and Claude Instant.

\bibitem{bal2025potential}
Mazhar Bal and Emre {\"O}zt{\"u}rk.
\newblock The potential of deep learning in improving k-12 students' writing skills: A systematic review.
\newblock {\em British Educational Research Journal}, 2025.

\bibitem{bandarkar2023belebele}
Lucas Bandarkar, Davis Liang, Benjamin Muller, Mikel Artetxe, Satya~Narayan Shukla, Donald Husa, Naman Goyal, Abhinandan Krishnan, Luke Zettlemoyer, and Madian Khabsa.
\newblock The belebele benchmark: a parallel reading comprehension dataset in 122 language variants.
\newblock In Lun{-}Wei Ku, Andre Martins, and Vivek Srikumar, editors, {\em Proceedings of the 62nd Annual Meeting of the Association for Computational Linguistics (Volume 1: Long Papers), {ACL} 2024, Bangkok, Thailand, August 11-16, 2024}, pages 749--775. Association for Computational Linguistics, 2024.

\bibitem{DBLP:journals/corr/abs-2504-05736}
Yida Cai, Kun Liang, Sanwoo Lee, Qinghan Wang, and Yunfang Wu.
\newblock Rank-then-score: Enhancing large language models for automated essay scoring.
\newblock {\em CoRR}, abs/2504.05736, 2025.

\bibitem{cevoli2022prediction}
Benedetta Cevoli, Chris Watkins, and Kathleen Rastle.
\newblock Prediction as a basis for skilled reading: Insights from modern language models.
\newblock {\em Royal Society open science}, 9(6):211837, 2022.

\bibitem{chardonnens2025adapting}
Sarah Chardonnens.
\newblock Adapting educational practices for generation z: integrating metacognitive strategies and artificial intelligence.
\newblock In {\em Frontiers in Education}, volume~10, page 1504726. Frontiers, 2025.

\bibitem{chen2021evaluating}
Mark Chen, Jerry Tworek, Heewoo Jun, Qiming Yuan, Henrique~Pond{\'{e}} de~Oliveira~Pinto, Jared Kaplan, Harri Edwards, Yuri Burda, Nicholas Joseph, Greg Brockman, Alex Ray, Raul Puri, Gretchen Krueger, Michael Petrov, Heidy Khlaaf, Girish Sastry, Pamela Mishkin, Brooke Chan, Scott Gray, Nick Ryder, Mikhail Pavlov, Alethea Power, Lukasz Kaiser, Mohammad Bavarian, Clemens Winter, Philippe Tillet, Felipe~Petroski Such, Dave Cummings, Matthias Plappert, Fotios Chantzis, Elizabeth Barnes, Ariel Herbert{-}Voss, William~Hebgen Guss, Alex Nichol, Alex Paino, Nikolas Tezak, Jie Tang, Igor Babuschkin, Suchir Balaji, Shantanu Jain, William Saunders, Christopher Hesse, Andrew~N. Carr, Jan Leike, Joshua Achiam, Vedant Misra, Evan Morikawa, Alec Radford, Matthew Knight, Miles Brundage, Mira Murati, Katie Mayer, Peter Welinder, Bob McGrew, Dario Amodei, Sam McCandlish, Ilya Sutskever, and Wojciech Zaremba.
\newblock Evaluating large language models trained on code.
\newblock {\em CoRR}, abs/2107.03374, 2021.

\bibitem{collins2005predicting}
Kevyn Collins-Thompson and Jamie Callan.
\newblock Predicting reading difficulty with statistical language models.
\newblock {\em Journal of the american society for information science and technology}, 56(13):1448--1462, 2005.

\bibitem{crossley2008assessing}
Scott~A Crossley, Jerry Greenfield, and Danielle~S McNamara.
\newblock Assessing text readability using cognitively based indices.
\newblock {\em Tesol Quarterly}, 42(3):475--493, 2008.

\bibitem{crossley2019moving}
Scott~A Crossley, Stephen Skalicky, and Mihai Dascalu.
\newblock Moving beyond classic readability formulas: New methods and new models.
\newblock {\em Journal of Research in Reading}, 42(3-4):541--561, 2019.

\bibitem{DBLP:journals/corr/abs-2412-19437}
DeepSeek{-}AI, Aixin Liu, Bei Feng, Bing Xue, Bingxuan Wang, Bochao Wu, Chengda Lu, Chenggang Zhao, Chengqi Deng, Chenyu Zhang, Chong Ruan, Damai Dai, Daya Guo, Dejian Yang, Deli Chen, Dongjie Ji, Erhang Li, Fangyun Lin, Fucong Dai, Fuli Luo, Guangbo Hao, Guanting Chen, Guowei Li, H.~Zhang, Han Bao, Hanwei Xu, Haocheng Wang, Haowei Zhang, Honghui Ding, Huajian Xin, Huazuo Gao, Hui Li, Hui Qu, J.~L. Cai, Jian Liang, Jianzhong Guo, Jiaqi Ni, Jiashi Li, Jiawei Wang, Jin Chen, Jingchang Chen, Jingyang Yuan, Junjie Qiu, Junlong Li, Junxiao Song, Kai Dong, Kai Hu, Kaige Gao, Kang Guan, Kexin Huang, Kuai Yu, Lean Wang, Lecong Zhang, Lei Xu, Leyi Xia, Liang Zhao, Litong Wang, Liyue Zhang, Meng Li, Miaojun Wang, Mingchuan Zhang, Minghua Zhang, Minghui Tang, Mingming Li, Ning Tian, Panpan Huang, Peiyi Wang, Peng Zhang, Qiancheng Wang, Qihao Zhu, Qinyu Chen, Qiushi Du, R.~J. Chen, R.~L. Jin, Ruiqi Ge, Ruisong Zhang, Ruizhe Pan, Runji Wang, Runxin Xu, Ruoyu Zhang, Ruyi Chen, S.~S. Li, Shanghao Lu, Shangyan Zhou,
  Shanhuang Chen, Shaoqing Wu, Shengfeng Ye, Shengfeng Ye, Shirong Ma, Shiyu Wang, Shuang Zhou, Shuiping Yu, Shunfeng Zhou, Shuting Pan, T.~Wang, Tao Yun, Tian Pei, Tianyu Sun, W.~L. Xiao, and Wangding Zeng.
\newblock Deepseek-v3 technical report.
\newblock {\em CoRR}, abs/2412.19437, 2024.

\bibitem{DBLP:journals/corr/abs-2505-23843}
Wenhan Dong, Tianyi Hu, Jingyi Zheng, Zhen Sun, Yuemeng Zhao, Yule Liu, Xinlei He, and Xinyi Huang.
\newblock Evaluation hallucination in multi-round incomplete information lateral-driven reasoning tasks.
\newblock {\em CoRR}, abs/2505.23843, 2025.

\bibitem{DBLP:journals/corr/abs-2505-00049}
Wenhan Dong, Yuemeng Zhao, Zhen Sun, Yule Liu, Zifan Peng, Jingyi Zheng, Zongmin Zhang, Ziyi Zhang, Jun Wu, Ruiming Wang, Shengmin Xu, Xinyi Huang, and Xinlei He.
\newblock Humanizing llms: {A} survey of psychological measurements with tools, datasets, and human-agent applications.
\newblock {\em CoRR}, abs/2505.00049, 2025.

\bibitem{ge2024chatgpt}
Hanjie Ge.
\newblock The role of {C}hat{GPT} in chinese reading education for chinese as a heritage language (chl) learners.
\newblock In {\em Proceedings of the 7th International Conference on Big Data and Education (ICBDE)}, 2024.

\bibitem{ghafouri2024virtual}
Mohammad Ghafouri, Jaleh Hassaskhah, and Amir Mahdavi-Zafarghandi.
\newblock From virtual assistant to writing mentor: Exploring the impact of a chatgpt-based writing instruction protocol on efl teachers’ self-efficacy and learners’ writing skill.
\newblock {\em Language Teaching Research}, page 13621688241239764, 2024.

\bibitem{goyal2022flores}
Naman Goyal, Cynthia Gao, Vishrav Chaudhary, Peng-Jen Chen, Guillaume Wenzek, Da~Ju, Sanjana Krishnan, Marc’Aurelio Ranzato, Francisco Guzm{\'a}n, and Angela Fan.
\newblock The flores-101 evaluation benchmark for low-resource and multilingual machine translation.
\newblock {\em Transactions of the Association for Computational Linguistics}, 10:522--538, 2022.

\bibitem{gu2025reading}
Chanyuan Gu, Samuel~A. Nastase, Zaid Zada, and Ping Li.
\newblock Reading comprehension in l1 and l2 readers: neurocomputational mechanisms revealed through large language models.
\newblock {\em NPJ Science of Learning}, 10(1):46, 2025.

\bibitem{hendrycks2020measuring}
Dan Hendrycks, Collin Burns, Steven Basart, Andy Zou, Mantas Mazeika, Dawn Song, and Jacob Steinhardt.
\newblock Measuring massive multitask language understanding.
\newblock {\em CoRR}, abs/2009.03300, 2020.

\bibitem{DBLP:conf/iclr/HuSWALWWC22}
Edward~J. Hu, Yelong Shen, Phillip Wallis, Zeyuan Allen{-}Zhu, Yuanzhi Li, Shean Wang, Lu~Wang, and Weizhu Chen.
\newblock Lora: Low-rank adaptation of large language models.
\newblock In {\em The Tenth International Conference on Learning Representations, {ICLR} 2022, Virtual Event, April 25-29, 2022}. OpenReview.net, 2022.

\bibitem{huynh2025genai}
Linh Huynh and Danielle~S. McNamara.
\newblock {GenAI}-powered text personalization: Natural language processing validation of adaptation capabilities.
\newblock {\em Applied Sciences}, 15(12):6791, 2025.

\bibitem{kim2023study}
Sunyoung Kim, Joobo Shim, Jaechang Shim, et~al.
\newblock A study on the utilization of openai chatgpt as a second language learning tool.
\newblock {\em Journal of Multimedia Information System}, 10(1):79--88, 2023.

\bibitem{li2023generaltextembeddingsmultistage}
Zehan Li, Xin Zhang, Yanzhao Zhang, Dingkun Long, Pengjun Xie, and Meishan Zhang.
\newblock Towards general text embeddings with multi-stage contrastive learning, 2023.

\bibitem{lopes2024language}
Adrielli~Tina Lopes~Rego, Joshua Snell, and Martijn Meeter.
\newblock Language models outperform cloze predictability in a cognitive model of reading.
\newblock {\em PLOS Computational Biology}, 20(9):e1012117, 2024.

\bibitem{LUI20161}
Angela~M. Lui and Sarah~M. Bonner.
\newblock Preservice and inservice teachers' knowledge, beliefs, and instructional planning in primary school mathematics.
\newblock {\em Teaching and Teacher Education}, 56:1--13, 2016.

\bibitem{ministry2022compulsory}
{Ministry of Education of the People’s Republic of China}.
\newblock Chinese curriculum standards for compulsory education, 2022.

\bibitem{ministry2022MathematicsCompulsory}
{Ministry of Education of the People’s Republic of China}.
\newblock Mathematics curriculum standards for compulsory education, 2022.

\bibitem{naz2024exploring}
Irum Naz and Rodney Robertson.
\newblock Exploring the feasibility and efficacy of chatgpt3 for personalized feedback in teaching.
\newblock {\em Electronic Journal of e-Learning}, 22(2):98--111, 2024.

\bibitem{nguyen2023efl}
Hang Nguyen Thi~Thu.
\newblock Efl teachers’ perspectives toward the use of chatgpt in writing classes: A case study at van lang university.
\newblock {\em Nguyen, TTH (2023). EFL Teachers’ Perspectives toward the Use of ChatGPT in Writing Classes: A Case Study at Van Lang University. International Journal of Language Instruction}, 2(3):1--47, 2023.

\bibitem{DBLP:journals/corr/abs-2303-08774}
OpenAI.
\newblock {GPT-4} technical report.
\newblock {\em CoRR}, abs/2303.08774, 2023.

\bibitem{opesemowo2024systematic}
Oluwaseyi Aina~Gbolade Opesemowo and Habeeb~Omoponle Adewuyi.
\newblock A systematic review of artificial intelligence in mathematics education: The emergence of 4ir.
\newblock {\em Eurasia Journal of Mathematics, Science and Technology Education}, 20(7):em2478, 2024.

\bibitem{rego2024cloze}
Adrielli T.~Lopes Rego, Joshua Snell, and Martijn Meeter.
\newblock Language models outperform cloze predictability in a cognitive model of reading.
\newblock {\em PLoS Computational Biology}, 20(9):e1012117, 2024.

\bibitem{sanghera2025high}
Rohan Sanghera, Arun~James Thirunavukarasu, Marc El~Khoury, Jessica O’Logbon, Yuqing Chen, Archie Watt, Mustafa Mahmood, Hamid Butt, George Nishimura, and Andrew~AS Soltan.
\newblock High-performance automated abstract screening with large language model ensembles.
\newblock {\em Journal of the American Medical Informatics Association}, 32(5):893--904, 2025.

\bibitem{Schneider01072013}
M.~Christina Schneider and Pamela Gowan.
\newblock Investigating teachers’ skills in interpreting evidence of student learning.
\newblock {\em Applied Measurement in Education}, 26(3):191--204, 2013.

\bibitem{vorobyeva2025personalized}
Klarisa~I Vorobyeva, Svetlana Belous, Natalia~V Savchenko, Lyudmila~M Smirnova, Svetlana~A Nikitina, and Sergei~P Zhdanov.
\newblock Personalized learning through ai: Pedagogical approaches and critical insights.
\newblock {\em Contemporary Educational Technology}, 17(2), 2025.

\bibitem{vygotsky1978mind}
Lev~Semenovich Vygotsky and Michael Cole.
\newblock {\em Mind in society: Development of higher psychological processes}.
\newblock Harvard university press, 1978.

\bibitem{DBLP:journals/corr/abs-2403-18105}
Shen Wang, Tianlong Xu, Hang Li, Chaoli Zhang, Joleen Liang, Jiliang Tang, Philip~S. Yu, and Qingsong Wen.
\newblock Large language models for education: {A} survey and outlook.
\newblock {\em CoRR}, abs/2403.18105, 2024.

\bibitem{wang2024mmlu}
Yubo Wang, Xueguang Ma, Ge~Zhang, Yuansheng Ni, Abhranil Chandra, Shiguang Guo, Weiming Ren, Aaran Arulraj, Xuan He, Ziyan Jiang, et~al.
\newblock Mmlu-pro: A more robust and challenging multi-task language understanding benchmark.
\newblock {\em Advances in Neural Information Processing Systems}, 37:95266--95290, 2024.

\bibitem{DBLP:conf/lak/XiaoMSXZWF25}
Changrong Xiao, Wenxing Ma, Qingping Song, Sean~Xin Xu, Kunpeng Zhang, Yufang Wang, and Qi~Fu.
\newblock Human-ai collaborative essay scoring: {A} dual-process framework with llms.
\newblock In {\em Proceedings of the 15th International Learning Analytics and Knowledge Conference, {LAK} 2025, Dublin, Ireland, March 3-7, 2025}, pages 293--305. {ACM}, 2025.

\bibitem{xiao2023chatgpt}
Changrong Xiao, Sean~Xin Xu, Kunpeng Zhang, Yufang Wang, and Lei Xia.
\newblock Evaluating reading comprehension exercises generated by {LLM}s: A showcase of {C}hat{GPT} in education applications.
\newblock In {\em Proceedings of the 18th Workshop on Innovative Use of NLP for Building Educational Applications (BEA 2023)}, 2023.

\bibitem{xiao2025exploring}
Feiwen Xiao, Siyu Zhu, and Wen Xin.
\newblock Exploring the landscape of generative ai (chatgpt)-powered writing instruction in english as a foreign language education: A scoping review.
\newblock {\em ECNU Review of Education}, page 20965311241310881, 2025.

\bibitem{DBLP:journals/corr/abs-2405-13001}
Hanyi Xu, Wensheng Gan, Zhenlian Qi, Jiayang Wu, and Philip~S. Yu.
\newblock Large language models for education: {A} survey.
\newblock {\em CoRR}, abs/2405.13001, 2024.

\bibitem{DBLP:journals/corr/abs-2306-01006}
Gautam Yadav.
\newblock Scaling evidence-based instructional design expertise through large language models.
\newblock {\em CoRR}, abs/2306.01006, 2023.

\bibitem{DBLP:journals/corr/abs-2412-15115}
An~Yang, Baosong Yang, Beichen Zhang, Binyuan Hui, Bo~Zheng, Bowen Yu, Chengyuan Li, Dayiheng Liu, Fei Huang, Haoran Wei, Huan Lin, Jian Yang, Jianhong Tu, Jianwei Zhang, Jianxin Yang, Jiaxi Yang, Jingren Zhou, Junyang Lin, Kai Dang, Keming Lu, Keqin Bao, Kexin Yang, Le~Yu, Mei Li, Mingfeng Xue, Pei Zhang, Qin Zhu, Rui Men, Runji Lin, Tianhao Li, Tingyu Xia, Xingzhang Ren, Xuancheng Ren, Yang Fan, Yang Su, Yichang Zhang, Yu~Wan, Yuqiong Liu, Zeyu Cui, Zhenru Zhang, and Zihan Qiu.
\newblock Qwen2.5 technical report.
\newblock {\em CoRR}, abs/2412.15115, 2024.

\bibitem{DBLP:journals/corr/abs-2504-05370}
Xueqiao Zhang, Chao Zhang, Jianwen Sun, Jun Xiao, Yi~Yang, and Yawei Luo.
\newblock Eduplanner: Llm-based multi-agent systems for customized and intelligent instructional design.
\newblock {\em CoRR}, abs/2504.05370, 2025.

\bibitem{zhipu_bigmodel_introduction}
{Zhipu AI}.
\newblock Introduction.
\newblock Zhipu AI Open Platform website (BigModel.cn), 2024.
\newblock Introduction of Zhipu AI Open Platform and the launch of GLM‑4 (January 16 2024).

\end{thebibliography}

\appendix
\newpage
 
\begin{table*}[t]
\setlength{\tabcolsep}{6pt} 
\renewcommand{\arraystretch}{0.95} 
\centering
\caption{Performance evaluation of 10 different models across various text genres in ZP setting: 
Results reveal that LLMs such as Qwen and GLM, which perform well on other leaderboards, exhibit poor performance in this task, even falling below the level of random guessing. 
This suggests that these LLMs may not have been exposed to relevant SCA tasks during their earlier training stages.}
\label{tab:big_zero}
\resizebox{0.98\textwidth}{!}{
\begin{tabular}{c|c|c|c|c|c|c|c|c|c|c|c}
\toprule
Dataset&\makecell{Model$\rightarrow$\\ Metrics $\downarrow$}& GPT4o &GPT4o-mini&R1&V3&Claude&GLM&Qwen-max&Qwen-plus&Qwen72B&Qwen32B\\
\midrule
\multirow{4}{*}{ALL}
&\makecell{F1} & 0.5318 & 0.6535 & 0.6851 &  {0.7933} & 0.4400 & 0.2480 & 0.2880 & 0.3048 & 0.4391 & 0.6696 \\
 
&\makecell{ACC} & 0.4942 & 0.5998 & 0.6835 &  {0.7775} & 0.4089 & 0.2659 & 0.3061 & 0.3152 & 0.4026 & 0.6314 \\
&\makecell{CCME} & 0.0605 & 0.0136 & 0.0075 & 0.0238 & 0.0135 &  0.0000 & 0.0079 & 0.0259 & 0.0159 & 0.0046 \\
&\makecell{WDBI} & 0.4796 & 0.1973 & -0.2220 & 0.1928 & 0.3720 &  {-0.0038} & 0.3452 & 0.4390 & 0.3393 & -0.1046 \\
\midrule
\multirow{4}{*}{Academic}
&\makecell{F1} &  {1.0000} & 0.9041 & 0.1026 & 0.8857 & 0.9610 & 0.1860 & 0.8358 & 0.9296 & 0.9459 & 0.4400 \\

&\makecell{ACC} &  {1.0000} & 0.8250 & 0.0541 & 0.7949 & 0.9250 & 0.1026 & 0.7179 & 0.8684 & 0.8974 & 0.2821 \\
&\makecell{CCME} & 0.0000 & 0.0000 & 0.0000 & 0.0000 & 0.0000 & 0.0000 & 0.0000 & 0.6000 & 0.0000 & 0.0000 \\
&\makecell{WDBI} &  0.0000  & -0.1750 & -0.9459 & -0.2051 & -0.0750 & -0.8974 & -0.2821 & -0.1316 & -0.1026 & -0.7179 \\
\midrule
\multirow{4}{*}{Adventure}
&\makecell{F1} & 0.2389 & 0.4105 & 0.5561 &  {0.7012} & 0.3593 & 0.3638 & 0.3721 & 0.3566 & 0.3434 & 0.4892 \\
 
&\makecell{ACC} & 0.2267 & 0.5067 & 0.6027 & {0.6933} & 0.5067 & 0.5270 & 0.5200 & 0.4800 & 0.4667 & 0.5867\\
&\makecell{CCME} & 0.3103 & 0.1351 & 0.0000 & 0.1739 & 0.0811 & 0.0000 & 0.1389 & 0.1538 & 0.1000 & 0.0000 \\
&\makecell{WDBI} & 0.7821 & 0.5107 & 0.3164 &  {0.2906} & 0.5128 & 0.5000 & 0.5000 & 0.5385 & 0.5513 & 0.4306 \\
\midrule
\multirow{4}{*}{Historical}
&\makecell{F1} & 0.4673 &  {0.7116} & 0.4152 & 0.6373 & 0.6607 & 0.4136 & 0.7195 & 0.5558 & 0.6480 & 0.6609 \\
 
&\makecell{ACC} & 0.5867 &  {0.7533} & 0.4815 & 0.6966 & 0.7133 & 0.4832 & 0.7551 & 0.6204 & 0.6939 & 0.6712 \\
&\makecell{CCME} & 0.0806 & 0.0000 & 0.0000 & 0.0000 & 0.0000 & 0.0000 & 0.0000 & 0.1154 & 0.0000 & 0.0000 \\
&\makecell{WDBI} & 0.6415 & 0.4535 &  {-0.0159} & 0.5321 & 0.5118 & 0.0763 & 0.4074 & 0.5262 & 0.4861 & 0.1614 \\
\midrule
\multirow{4}{*}{Philosophy}
&\makecell{F1} & 0.8521 & 0.9114 & 0.4413 &  {0.9747} & 0.9615 & 0.4070 & 0.8813 & 0.8862 & 0.8654 & 0.7897 \\
 
&\makecell{ACC} & 0.8929 & 0.9048 & 0.3855 &  {0.9759} & 0.9643 & 0.3571 & 0.8675 & 0.8765 & 0.8554 & 0.7470 \\
&\makecell{CCME} & 0.0000 & 0.0000 & 0.0000 & 0.0000 & 0.0000 & 0.0000 & 0.0000 & 0.4000 & 0.0000 & 0.0000 \\
&\makecell{WDBI} & 0.4500 & 0.0595 & -0.3493 & 0.1000 & 0.1500 & -0.3649 &  {0.0384} & 0.2718 & 0.1452 & -0.0870 \\
\midrule
\multirow{4}{*}{\makecell{Family \\ and \\ Campus}}
&\makecell{F1} & 0.5901 & 0.7781 &  {0.9158} & 0.9112 & 0.4688 & 0.3172 & 0.2415 & 0.1949 & 0.4528 & 0.8201 \\
 
&\makecell{ACC} & 0.4734 & 0.7174 &  {0.9031} & 0.8913 & 0.3768 & 0.2705 & 0.2174 & 0.1800 & 0.3671 & 0.7754 \\
&\makecell{CCME} & 0.0046 & 0.0000 & 0.0000 & 0.0000 & 0.0039 & 0.0000 & 0.0000 & 0.0000 & 0.0000 & 0.0000 \\
&\makecell{WDBI} & 0.4791 & 0.1674 &  {0.0099} & 0.0992 & 0.4141 & 0.4016 & 0.4427 & 0.5108 & 0.3957 & 0.0512 \\
\midrule
\multirow{4}{*}{Fantasy}
&\makecell{F1} & 0.1295 & 0.3004 & 0.7713 &  {0.7813} & 0.0233 & 0.0158 & 0.0052 & 0.0598 & 0.1931 & 0.7501 \\
 
&\makecell{ACC} & 0.0822 & 0.2283 & 0.6895 &  {0.6941} & 0.0548 & 0.0639 & 0.0457 & 0.0783 & 0.1370 & 0.6621 \\
&\makecell{CCME} & 0.0249 & 0.0118 & 0.0000 & 0.0000 & 0.0000 & 0.0000 & 0.0096 & 0.0020 & 0.0212 & 0.0000 \\
&\makecell{WDBI} & 0.8482 & 0.4462 &  {0.0833} & 0.2707 & 0.6105 & 0.4976 & 0.6154 & 0.5622 & 0.6749 & 0.1796 \\
\midrule
\multirow{4}{*}{\makecell{Literature \\ Fiction}}
&\makecell{F1} & 0.5368 &  {0.6596} & 0.5590 & 0.6572 & 0.6531 & 0.5393 & 0.6117 & 0.4970 & 0.6294 & 0.5230 \\
 
&\makecell{ACC} & 0.6166 & 0.6601 & 0.5878 &  {0.6825} & 0.6706 & 0.5754 & 0.6175 & 0.5344 & 0.6349 & 0.5516 \\
&\makecell{CCME} & 0.0619 & 0.0116 & 0.0000 & 0.0375 & 0.0120 & 0.0000 & 0.0000 & 0.1304 & 0.0109 & 0.0000 \\
&\makecell{WDBI} & 0.3808 & 0.1723 & -0.1593 & 0.2623 & 0.2891 &  {-0.0878} & 0.2898 & 0.3762 & 0.2471 & -0.1296 \\
\midrule
\multirow{4}{*}{Societal}
&\makecell{F1} & 0.6343 & 0.6228 & 0.3390 &  {0.6833} & 0.6015 & 0.2290 & 0.5476 & 0.5781 & 0.6022 & 0.4665 \\
 
&\makecell{ACC} & 0.6747 & 0.5901 & 0.3747 &  {0.6843} & 0.5840 & 0.2928 & 0.5288 & 0.5906 & 0.5744 & 0.4564 \\
&\makecell{CCME} & 0.0112 & 0.0088 & 0.0200 & 0.0039 & 0.0029 & 0.0000 & 0.0026 & 0.0364 & 0.0029 & 0.0181 \\
&\makecell{WDBI} & 0.4958 & 0.1801 & -0.1574 & 0.2871 & 0.3467 &  {-0.0119} & 0.3236 & 0.4341 & 0.2680 & -0.0617 \\
\midrule
\multirow{4}{*}{\makecell{Science \\ Fiction}}
&\makecell{F1} & 0.5655 & 0.5769 & 0.2879 &  {0.8272} & 0.5008 & 0.0090 & 0.3781 & 0.5209 & 0.4607 & 0.2501 \\
 
&\makecell{ACC} & 0.5949 & 0.4937 & 0.3026 &  {0.8101} & 0.4937 & 0.0633 & 0.3165 & 0.5325 & 0.4304 & 0.2025 \\
&\makecell{CCME} & 0.0312 & 0.0000 & 0.0000 & 0.0000 & 0.0000 & 0.0000 & 0.0000 & 0.0278 & 0.0000 & 0.0000 \\
&\makecell{WDBI} & 0.6543 & 0.4356 & -0.1844 & 0.3661 & 0.5531 & 0.0556 & 0.5145 & 0.6370 & 0.5169 &  {0.0454} \\
\midrule
\multirow{4}{*}{\makecell{Fairy \\ Tales}}
&\makecell{F1} & 0.6302 & 0.7684 & 0.9351 &  {0.9469} & 0.4029 & 0.2680 & 0.1383 & 0.1960 & 0.4469 & 0.8581 \\
 
&\makecell{ACC} & 0.4659 & 0.6393 & 0.8971 &  {0.9158} & 0.2635 & 0.1679 & 0.0861 & 0.1162 & 0.2992 & 0.7701 \\
&\makecell{CCME} & 0.0598 & 0.0122 & 0.0000 & 0.0224 & 0.0119 & 0.0000 & 0.0021 & 0.0086 & 0.0108 & 0.0000 \\
&\makecell{WDBI} & 0.6465 & 0.2532 &  {0.0280} & 0.1366 & 0.4671 & 0.4216 & 0.5100 & 0.6592 & 0.4491 & 0.1165 \\
\midrule
\multirow{4}{*}{Non-fiction}
&\makecell{F1} & 0.3876 & 0.5449 & 0.6005 &  {0.6457} & 0.3800 & 0.2195 & 0.2629 & 0.2759 & 0.2826 & 0.5432 \\
 
&\makecell{ACC} & 0.4086 & 0.5231 & 0.6186 &  {0.6301} & 0.3966 & 0.3148 & 0.3395 & 0.3278 & 0.3337 & 0.5367 \\
&\makecell{CCME} & 0.0917 & 0.0126 & 0.0027 & 0.0360 & 0.0250 & 0.0000 & 0.0232 & 0.0292 & 0.0337 & 0.0000 \\
&\makecell{WDBI} & 0.5483 & 0.2951 & -0.1964 & 0.2604 & 0.4212 &  {-0.0240} & 0.3980 & 0.4779 & 0.4648 & -0.0621 \\
\midrule
\multirow{4}{*}{Poetry}
&\makecell{F1} & 0.4497 & 0.6122 &  {0.9752} & 0.6870 & 0.4497 & 0.5684 & 0.5226 & 0.3343 & 0.4925 & 0.7297 \\
 
&\makecell{ACC} & 0.3415 & 0.5122 &  {0.9750} & 0.6000 & 0.3415 & 0.5854 & 0.4500 & 0.2250 & 0.4000 & 0.7250 \\
&\makecell{CCME} & 0.0741 & 0.1000 & 0.0000 & 0.0625 & 0.1111 & 0.0000 & 0.0000 & 0.0323 & 0.0417 & 0.0000 \\
&\makecell{WDBI} & 0.7407 & 0.5595 &  {0.0192} & 0.4725 & 0.7407 & 0.3148 & 0.5385 & 0.8269 & 0.6099 & 0.1566 \\
\bottomrule
\end{tabular}}
\end{table*}

\begin{table*}
\setlength{\tabcolsep}{6pt} 
\renewcommand{\arraystretch}{0.95} 
\centering
\caption{Performance evaluation of 10 different models across various text genres in CL setting.
``All'' represents the aggregate performance across all genres. 
F1 and ACC indicate the model's performance for a specific genre. CLME assesses the proportion of misclassifications occurring across non-adjacent difficulty levels, while WDBI evaluates the overall tendency of the model to overestimate or underestimate difficulty levels, accounting for class imbalance.}
\label{tab:big}
\resizebox{0.98\textwidth}{!}{
\begin{tabular}{c|c|c|c|c|c|c|c|c|c|c|c}
\toprule
Dataset&\makecell{Model$\rightarrow$\\ Metrics $\downarrow$}& GPT4o &GPT4o-mini&R1&V3&Claude&GLM&Qwen-max&Qwen-plus&Qwen72B&Qwen32B\\
\midrule
\multirow{4}{*}{All}
&\makecell{F1} & 0.6896 & 0.7993 & 0.7184 &  {0.8352} & 0.6860 & 0.7038 & 0.6818 & 0.6962 & 0.6547 & 0.7610 \\
 
&\makecell{ACC} & 0.6618 & 0.7784 & 0.7413 &  {0.8313} & 0.6445 & 0.6878 & 0.6296 & 0.6674 & 0.6091 & 0.7555 \\
&\makecell{CLME} & 0.0477 & 0.0189 & 0.0874 & 0.0350 & 0.0329 & 0.0074 &  {0.0062} & 0.0924 & 0.0361 & 0.0396 \\
&\makecell{WDBI} & 0.3561 &  {0.0704} & -0.3243 & 0.1081 & 0.2579 & -0.1708 & 0.3727 & 0.1684 & 0.2953 & -0.1930 \\
\midrule
\multirow{4}{*}{Academic}
&\makecell{F1} &  {0.9873} & 0.7500 & 0.2222 & 0.8571 & 0.9744 & 0.3750 & 0.7302 & 0.9333 & 0.9474 & 0.5185 \\

&\makecell{ACC} &  {0.9750} & 0.6000 & 0.1250 & 0.7500 & 0.9500 & 0.2308 & 0.5750 & 0.8750 & 0.9000 & 0.3500 \\
&\makecell{CLME} & 0.0000 & 0.0000 & 0.0000 & 0.0000 & 0.0000 & 0.0000 & 0.0000 & 0.6000 & 0.0000 & 0.0000 \\
&\makecell{WDBI} &  {-0.0250} & -0.4000 & -0.8750 & -0.2500 & -0.0500 & -0.7692 & 0.4250 & -0.1250 & -0.1000 & -0.6500 \\
\midrule
\multirow{4}{*}{Adventure}
&\makecell{F1} & 0.2859 & 0.6039 & 0.6000 &  {0.7672} & 0.4496 & 0.4989 & 0.3914 & 0.5114 & 0.3290 & 0.5867 \\
 
&\makecell{ACC} & 0.3067 & 0.6400 & 0.6027 & { 0.7467} & 0.5067 & 0.5946 & 0.5333 & 0.5479 & 0.4133 & 0.6000 \\
&\makecell{CLME} & 0.1538 & 0.1111 & 0.0000 & 0.2105 & 0.1351 & 0.0000 & 0.0571 & 0.1515 & 0.1818 & 0.0000 \\
&\makecell{WDBI} & 0.7041 & 0.3739 &  {-0.0965} & 0.2350 & 0.5085 & 0.4286 & 0.4933 & 0.3605 & 0.6026 & 0.1763 \\
\midrule
\multirow{4}{*}{Historical}
&\makecell{F1} & 0.5286 &  {0.7861} & 0.5473 & 0.7161 & 0.6190 & 0.5470 & 0.6976 & 0.5760 & 0.6335 & 0.7118 \\
 
&\makecell{ACC} & 0.6200 &  {0.8000} & 0.5532 & 0.7466 & 0.6824 & 0.5772 & 0.7230 & 0.6084 & 0.6913 & 0.7095 \\
&\makecell{CLME} & 0.0000 & 0.0000 & 0.0159 & 0.0000 & 0.0000 & 0.0000 & 0.0000 & 0.1429 & 0.0000 & 0.0000 \\
&\makecell{WDBI} & 0.6101 & 0.3001 & -0.2618 & 0.4103 & 0.5472 & 0.1634 & 0.2770 & 0.3623 & 0.5306 &  {0.1271} \\
\midrule
\multirow{4}{*}{Philosophy}
&\makecell{F1} & 0.9147 & 0.9394 & 0.5247 &  {0.9762} & 0.9147 & 0.6177 & 0.8403 & 0.8989 & 0.9184 & 0.7352 \\
 
&\makecell{ACC} & 0.9286 & 0.9286 & 0.4048 &  {0.9762} & 0.9286 & 0.5476 & 0.8095 & 0.8537 & 0.9167 & 0.6786 \\
&\makecell{CLME} & 0.0000 & 0.0000 & 0.0000 & 0.0000 & 0.0000 & 0.0000 & 0.0000 & 0.5833 & 0.0000 & 0.0000 \\
&\makecell{WDBI} & 0.3000 & -0.0838 & -0.5973 &  {0.0432} & 0.3000 & -0.2568 & 0.1905 & -0.1098 & 0.1230 & -0.1257 \\
\midrule
\multirow{4}{*}{\makecell{Family \\ and \\ Campus}}
&\makecell{F1} & 0.8000 & 0.9072 & 0.9258 &  {0.9563} & 0.7555 & 0.9058 & 0.7974 & 0.8494 & 0.7782 & 0.9200 \\
 
&\makecell{ACC} & 0.7391 & 0.8913 & 0.9396 &  {0.9588} & 0.6880 & 0.8913 & 0.7440 & 0.8100 & 0.7150 & 0.9275 \\
&\makecell{CLME} & 0.0093 & 0.0000 & 0.0000 & 0.0000 & 0.0000 & 0.0000 & 0.0000 & 0.0000 & 0.0000 & 0.0000 \\
&\makecell{WDBI} & 0.2146 & 0.0309 & -0.3289 & -0.1608 & 0.1812 &  {-0.0126} & 0.2560 & 0.0662 & 0.1634 & -0.2788 \\
\midrule
\multirow{4}{*}{Fantasy}
&\makecell{F1} & 0.4273 & 0.7598 & 0.9294 & 0.9166 & 0.4379 & 0.8748 & 0.5911 & 0.5993 & 0.4745 &  {0.9427} \\
 
&\makecell{ACC} & 0.2922 & 0.6712 & 0.9296 & 0.8904 & 0.3178 & 0.8356 & 0.4749 & 0.4717 & 0.3288 &  {0.9315} \\
&\makecell{CLME} & 0.0258 & 0.0139 & 0.0000 & 0.0417 & 0.0068 & 0.0000 & 0.0087 & 0.0268 & 0.0408 & 0.0000 \\
&\makecell{WDBI} & 0.7005 & 0.2108 & -0.2474 & 0.1303 & 0.5431 & 0.0465 & 0.5297 & 0.3842 & 0.7171 &  {0.0364} \\
\midrule
\multirow{4}{*}{\makecell{Literature \\ Fiction}}
&\makecell{F1} & 0.5698 & 0.6385 & 0.5575 & 0.6548 & 0.6313 & 0.6362 &  {0.6796} & 0.5781 & 0.6093 & 0.6166 \\
 
&\makecell{ACC} & 0.6245 & 0.6403 & 0.5628 &  {0.6759} & 0.6548 & 0.6508 & 0.6746 & 0.6082 & 0.6324 & 0.6206 \\
&\makecell{CLME} & 0.0632 & 0.0110 & 0.0741 & 0.0366 & 0.0575 & 0.0114 & 0.0000 & 0.2083 & 0.0108 & 0.1146 \\
&\makecell{WDBI} & 0.3435 & 0.0915 & -0.2010 & 0.2089 & 0.2226 & -0.1106 & 0.3254 & 0.1550 & 0.2624 &  {-0.0905} \\
\midrule
\multirow{4}{*}{Societal}
&\makecell{F1} & 0.6602 & 0.6629 & 0.3437 &  {0.7051} & 0.6550 & 0.4268 & 0.5686 & 0.6587 & 0.6665 & 0.4815 \\
 
&\makecell{ACC} & 0.6892 & 0.6397 & 0.3903 &  {0.7045} & 0.6407 & 0.4399 & 0.5324 & 0.6675 & 0.6504 & 0.4817 \\
&\makecell{CLME} & 0.0039 & 0.0168 & 0.1452 & 0.0372 & 0.0103 & 0.0217 & 0.0000 & 0.1742 & 0.0035 & 0.0824 \\
&\makecell{WDBI} & 0.4262 &  {0.1055} & -0.2986 & 0.1298 & 0.2528 & -0.1627 & 0.4676 & 0.1821 & 0.2698 & -0.1989 \\
\midrule
\multirow{4}{*}{\makecell{Science \\ Fiction}}
&\makecell{F1} & 0.6285 & 0.7335 & 0.3640 &  {0.8533} & 0.6297 & 0.2793 & 0.3800 & 0.6811 & 0.5391 & 0.4276 \\
 
&\makecell{ACC} & 0.6203 & 0.6582 & 0.4103 &  {0.8590} & 0.6234 & 0.2152 & 0.3038 & 0.6622 & 0.5385 & 0.3924 \\
&\makecell{CLME} & 0.0000 & 0.0000 & 0.0652 & 0.0000 & 0.0000 & 0.0000 & 0.0000 & 0.0000 & 0.0278 & 0.0000 \\
&\makecell{WDBI} & 0.6100 & 0.3339 & -0.2148 & 0.2971 & 0.5542 &  {0.0722} & 0.6962 & 0.5898 & 0.6176 & -0.1243 \\
\midrule
\multirow{4}{*}{\makecell{Fairy \\ Tales}}
&\makecell{F1} & 0.8537 & 0.9465 &  {0.9845} & 0.9731 & 0.7890 & 0.9042 & 0.8127 & 0.8556 & 0.7928 & 0.9647 \\
 
&\makecell{ACC} & 0.7564 & 0.9161 &  {0.9830} & 0.9624 & 0.6650 & 0.8443 & 0.7013 & 0.7636 & 0.6692 & 0.9499 \\
&\makecell{CLME} & 0.0643 & 0.0224 & 0.0000 & 0.0333 & 0.0341 & 0.0000 & 0.0021 & 0.0187 & 0.0322 & 0.0000 \\
&\makecell{WDBI} & 0.4288 & 0.0412 & -0.2085 & 0.0165 & 0.2859 &  {0.0065} & 0.2993 & 0.2131 & 0.3790 & -0.1435 \\
\midrule
\multirow{4}{*}{Non-fiction}
&\makecell{F1} & 0.5531 & 0.7179 & 0.6286 &  {0.7210} & 0.6541 & 0.6126 & 0.6030 & 0.5178 & 0.4644 & 0.6643 \\
 
&\makecell{ACC} & 0.5562 & 0.7028 & 0.6701 &  {0.7107} & 0.6442 & 0.6242 & 0.5777 & 0.5048 & 0.4671 & 0.6626 \\
&\makecell{CLME} & 0.0679 & 0.0203 & 0.0723 & 0.0281 & 0.0544 & 0.0000 & 0.0168 & 0.0901 & 0.0551 & 0.0000 \\
&\makecell{WDBI} & 0.4198 &  {0.1241} & -0.3182 & 0.1614 & 0.2679 & -0.1713 & 0.4294 & 0.2373 & 0.4520 & -0.1584 \\
\midrule
\multirow{4}{*}{Poetry}
&\makecell{F1} & 0.5268 & 0.6950 &  {0.8685} & 0.7647 & 0.5152 & 0.7365 & 0.6445 & 0.5670 & 0.5365 & 0.7500 \\
 
&\makecell{ACC} & 0.4390 & 0.6098 &  {0.8250} & 0.7000 & 0.4103 & 0.7073 & 0.5750 & 0.4857 & 0.4250 & 0.7500 \\
&\makecell{CLME} & 0.0870 & 0.0625 & 0.0000 & 0.0833 & 0.1739 & 0.0000 & 0.0000 & 0.0556 & 0.1739 & 0.0000 \\
&\makecell{WDBI} & 0.6667 & 0.4683 & 0.1621 & 0.3077 & 0.7037 & 0.2566 & 0.4250 & 0.5052 & 0.6401 &  {-0.0824} \\
\bottomrule
\end{tabular}}
\end{table*}

\newpage
\begin{figure*}
\centering
\captionsetup[subfigure]{labelformat=empty} 
\subfloat[GPT-4o]{\includegraphics[width=0.3\textwidth]{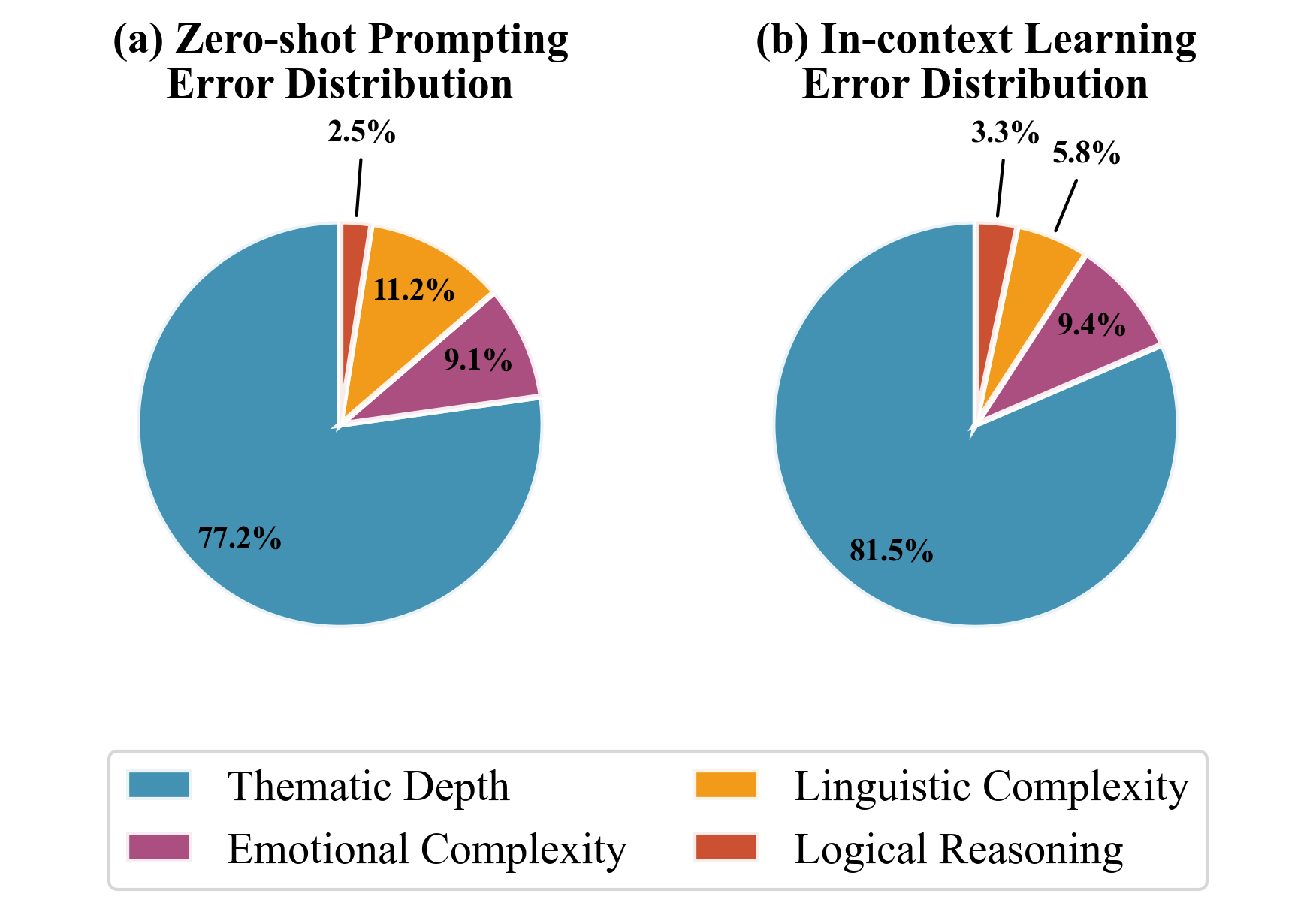}}\hfill
\subfloat[GPT-4o-mini]{\includegraphics[width=0.3\textwidth]{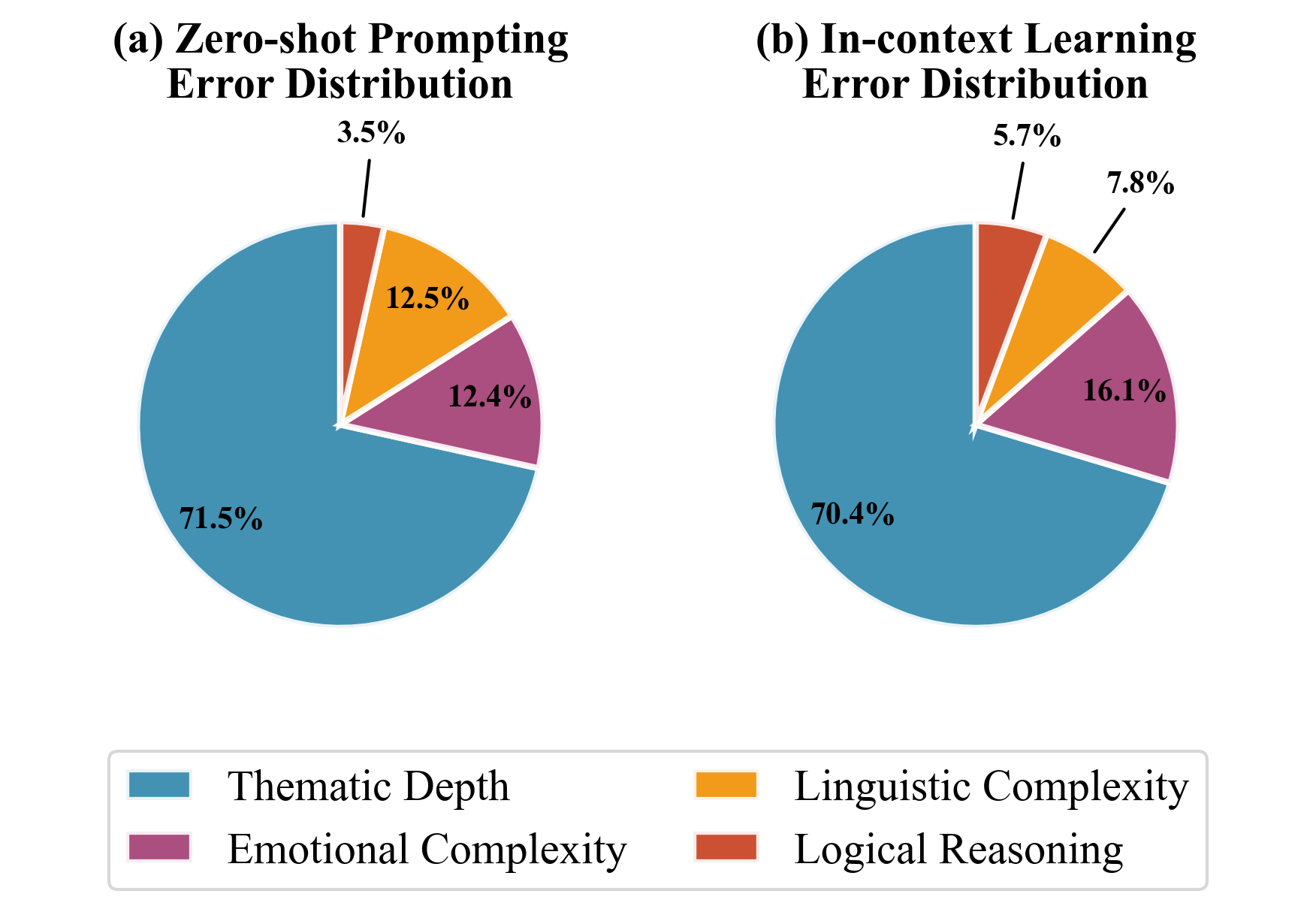}}\hfill
\subfloat[R1]{\includegraphics[width=0.3\textwidth]{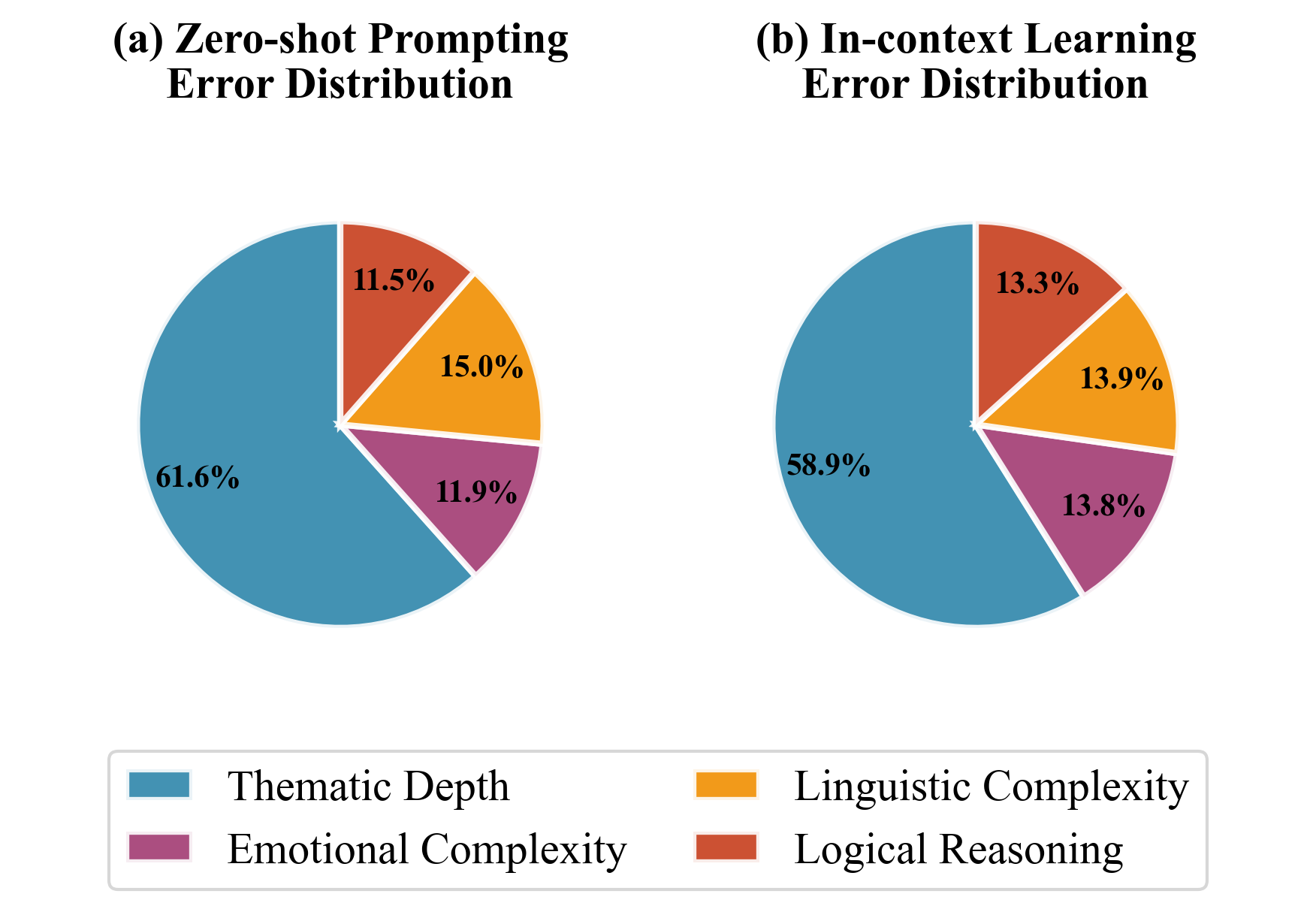}}
\vspace{0.4em}
\subfloat[V3]{\includegraphics[width=0.3\textwidth]{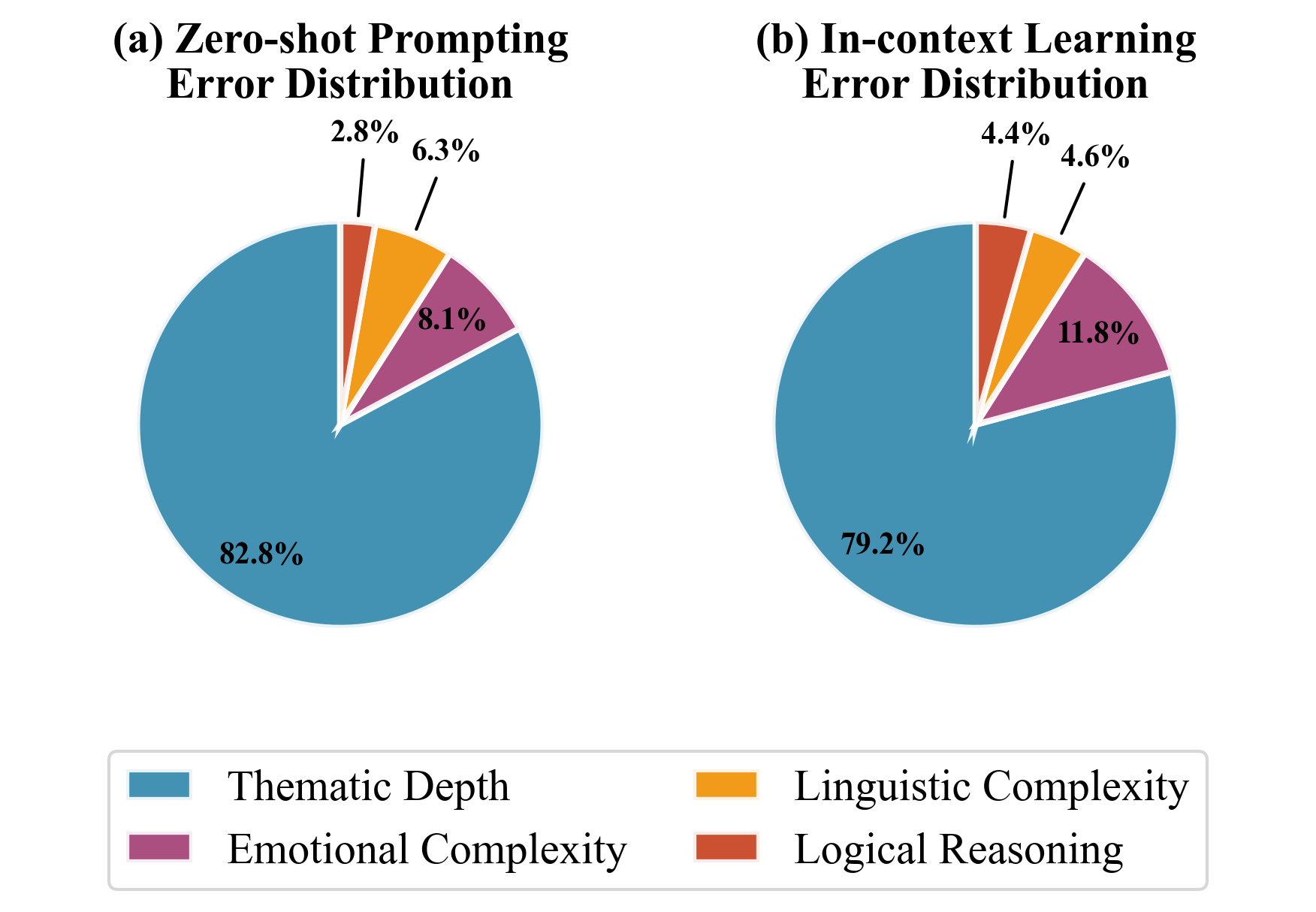}}\hfill
\subfloat[Claude]{\includegraphics[width=0.3\textwidth]{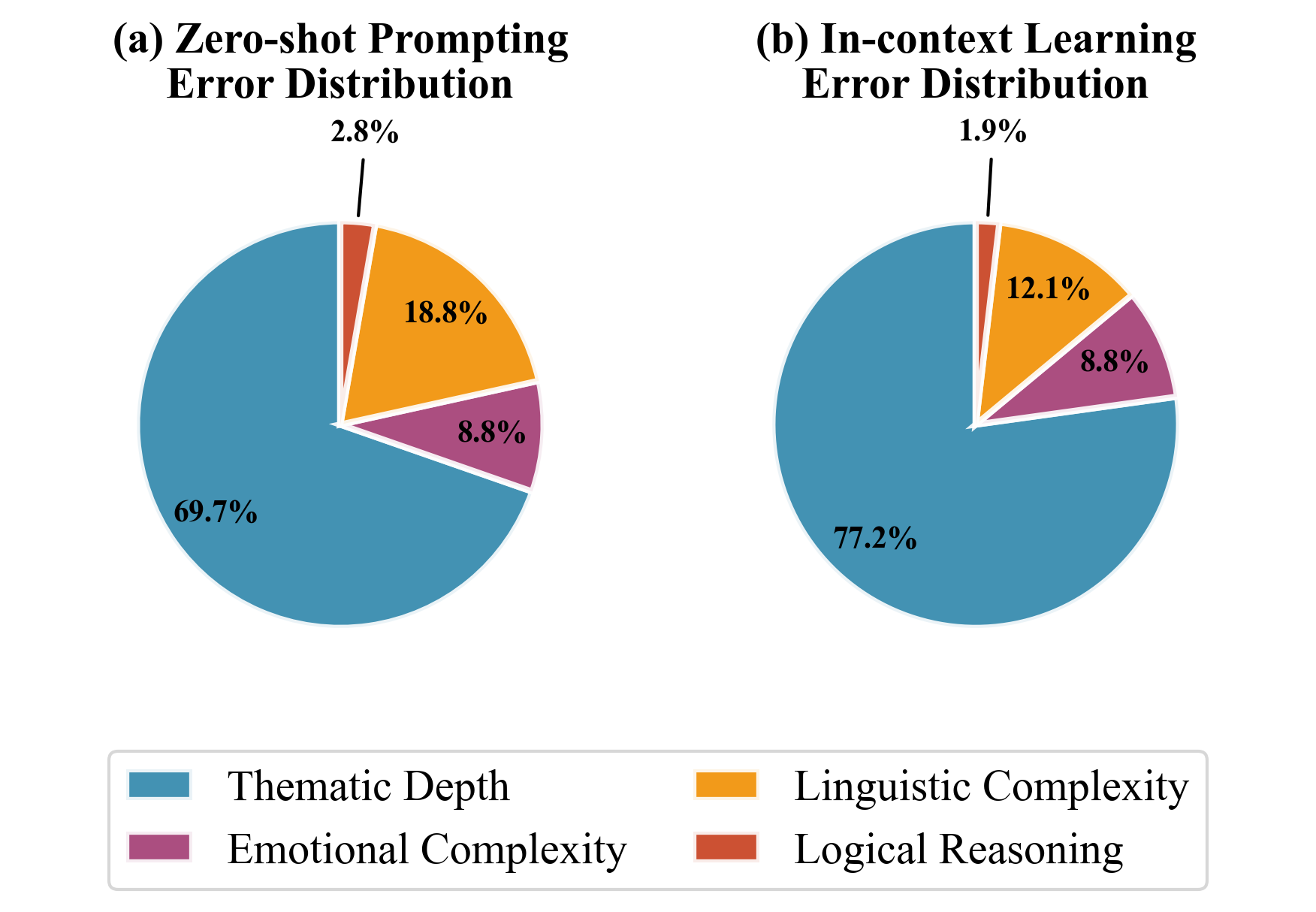}}\hfill
\subfloat[GLM]{\includegraphics[width=0.3\textwidth]{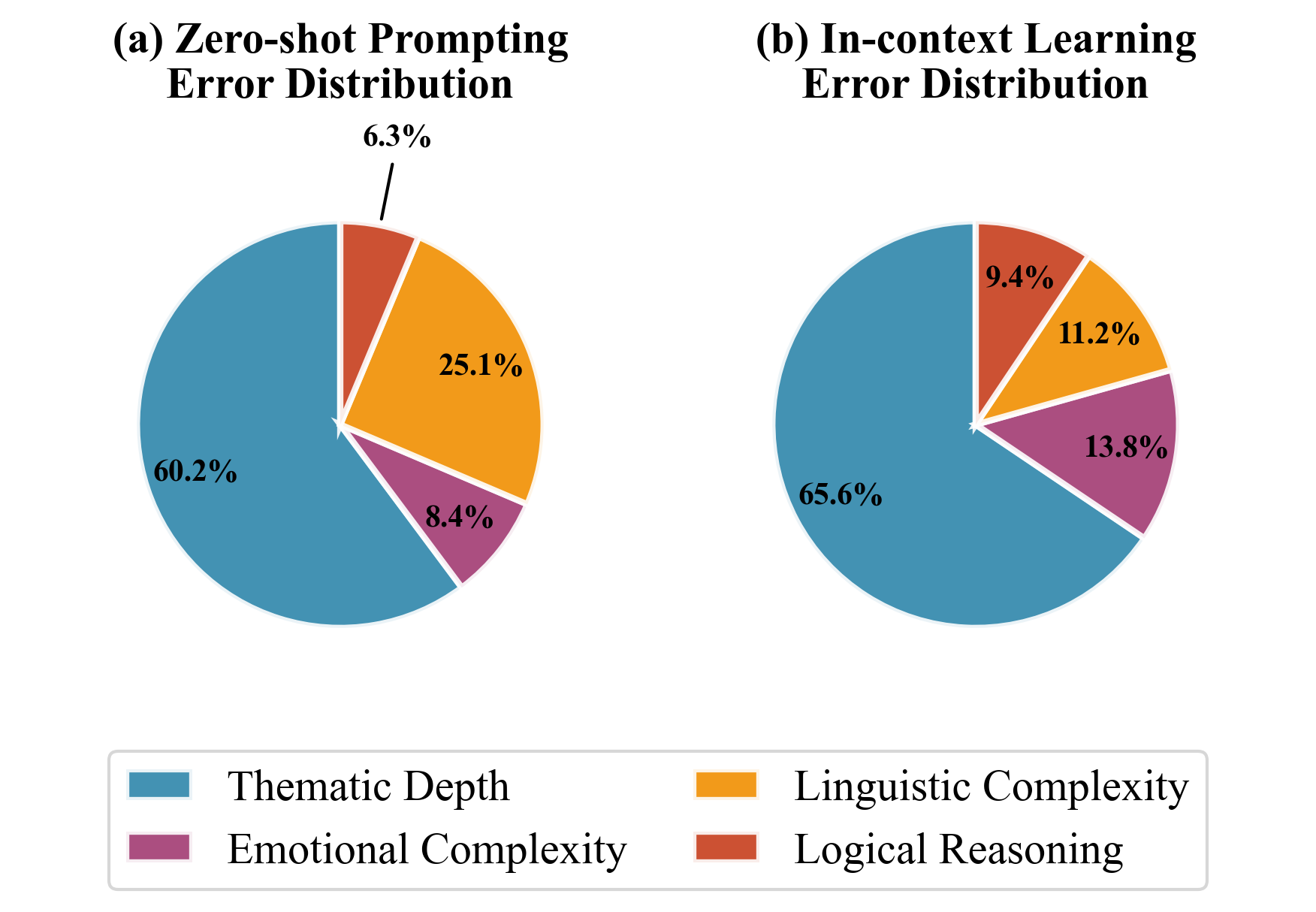}}
\vspace{0.4em}
\subfloat[Qwen-max]{\includegraphics[width=0.3\textwidth]{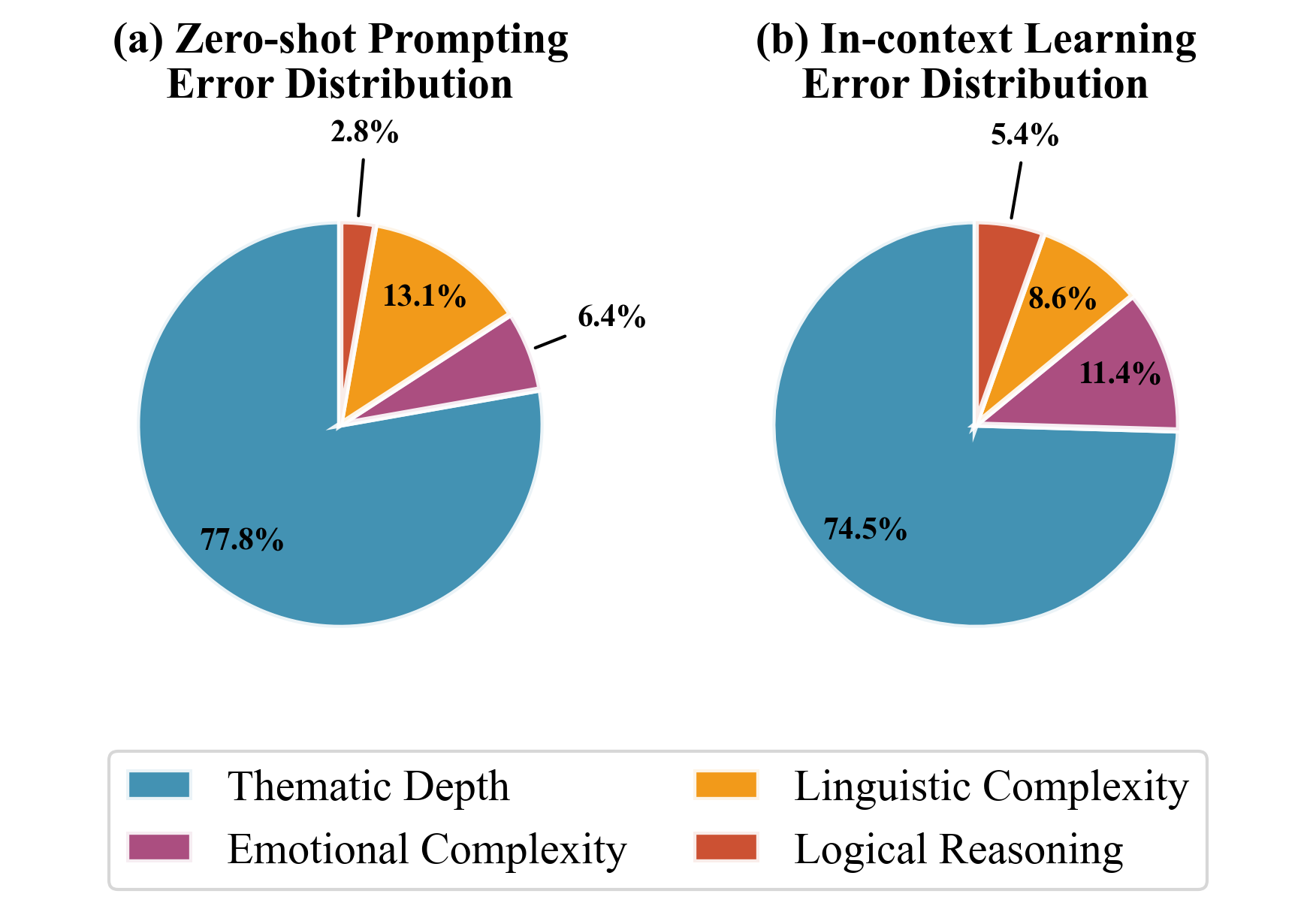}}\hfill
\subfloat[Qwen-plus]{\includegraphics[width=0.3\textwidth]{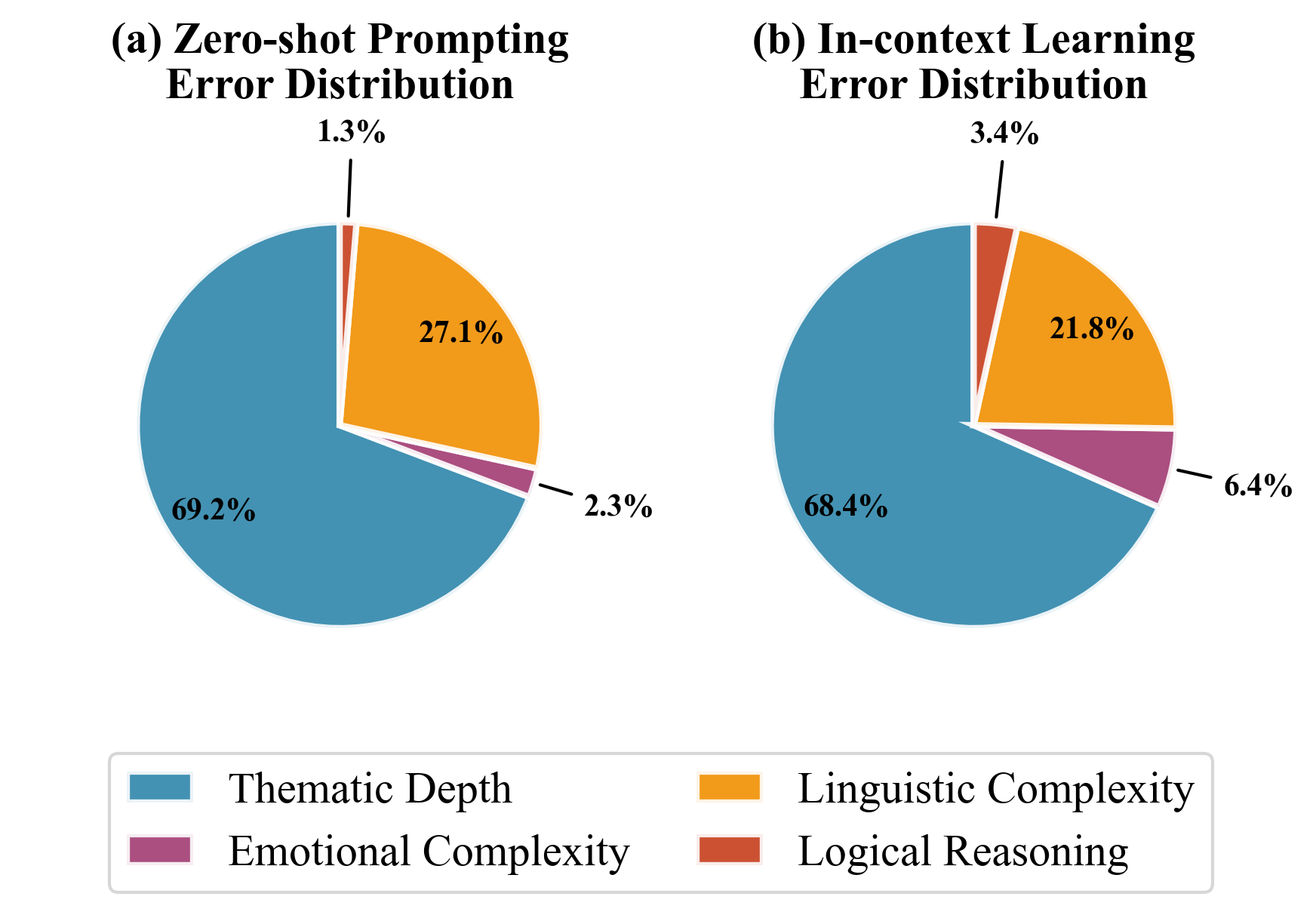}}\hfill
\subfloat[Qwen72B]{\includegraphics[width=0.3\textwidth]{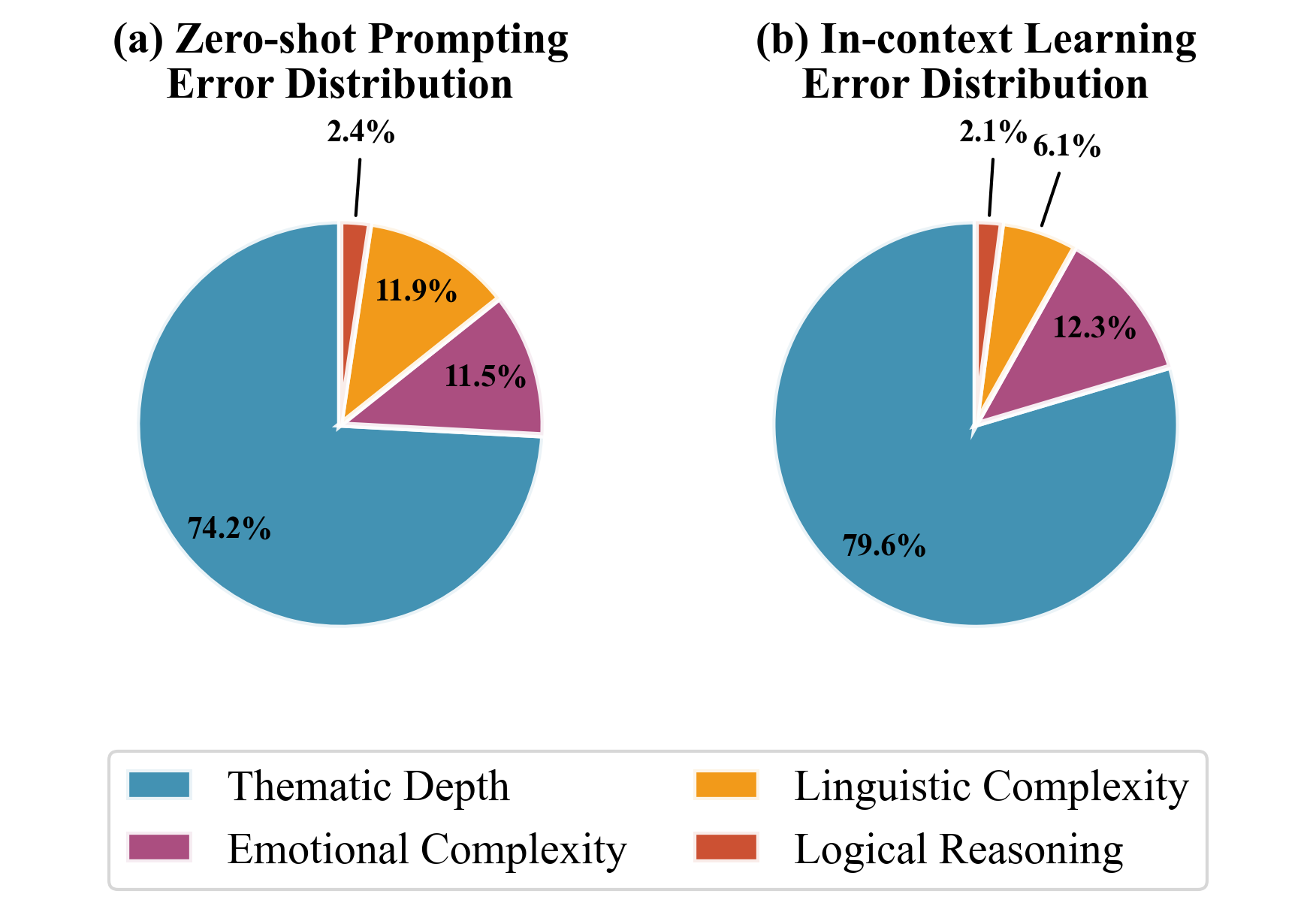}}

\vspace{0.4em}
\subfloat[Qwen32B]{\includegraphics[width=0.3\textwidth]{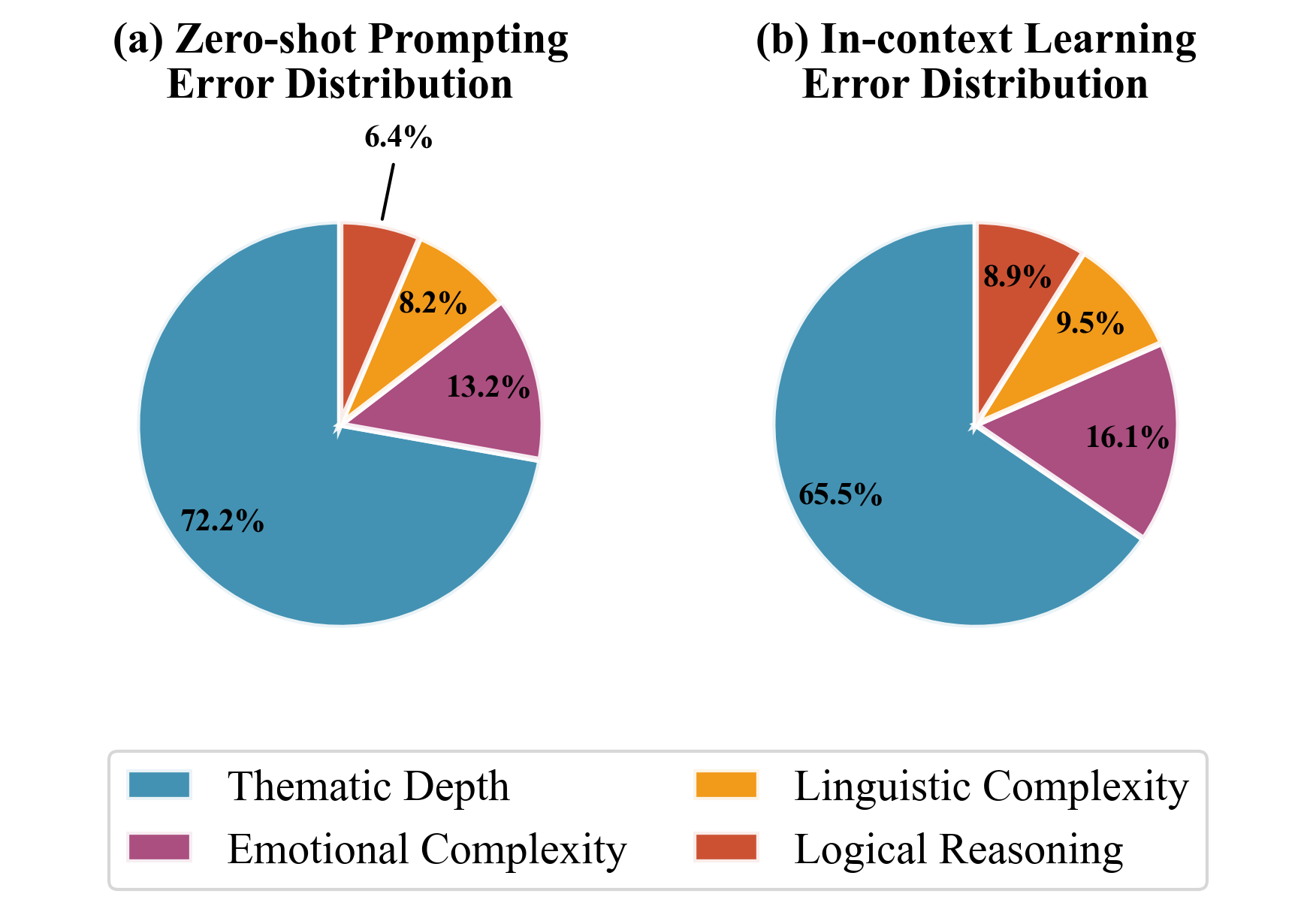}}
\caption{Comparative analysis of error type classification: Major errors in LLMs primarily stem from thematic depth. 
This may suggest that LLMs' misunderstanding of SCA could be attributed to ambiguities in recognizing the appropriate thematic for students at different educational stages.}
\label{fig:error_analysis}
\end{figure*}
\end{document}